# Protected QL-Memory: Diagnosing Capability Preservation and Task Sensitivity in Memory-Augmented Document Classifiers

Isaac Kofi Nti[1,2,*]
[1] School of Information Technology, University of Cincinnati, Cincinnati, OH, USA
[2] Information Technology Analytics Center, University of Cincinnati, Cincinnati, OH, USA
***Corresponding author:** Isaac Kofi Nti
**Email:** ntious1@gmail.com
**ORCID:** https://orcid.org/0000-0001-9257-4295

**Abstract**

**Purpose:** End-task accuracy alone cannot determine whether a memory mechanism preserves an acquired capability, exposes sample-specific stored information, or contributes measurably to downstream performance. This study introduces Protected QL-Memory and evaluates capability preservation, diagnostic access, and task-performance sensitivity as distinct empirical properties.

**Methods:** Protected QL-Memory is a dual-path document classifier combining a causal local pathway, an associative matrix writer, and a finalized-memory reader. A capability-protected schedule acquires a controlled binding capability, adapts the local pathway while constraining writer degradation, trains controlled memory access, and restricts full-path task fitting. Pre- and post-adaptation diagnostics and finalized-memory interventions were evaluated across nine dataset-seed conditions.

**Results:** Writer capability was fully preserved, with 100% post-adaptation accuracy. Cyclic reassignment of finalized matrices produced diagnostic accuracy gaps of 86.3–86.9 percentage points, showing strong dependence on example-memory correspondence. In contrast, natural-text macro-F1 changed by less than 0.002 when finalized matrices were reassigned, zeroed, or replaced by batch means. Locked-test macro-F1 was 90.94% on AG News, 81.78% on IMDB, and 62.63% on Yelp Review Full, comparable to compact controls.

**Conclusion:** Preserving associative capability and maintaining diagnostic access to sample-specific memory did not imply measurable downstream reliance on that memory. Protected-memory designs should therefore treat capability preservation, diagnostic access, and task alignment as separate objectives and evaluate downstream memory sensitivity directly.

## 1 INTRODUCTION

Document decisions are often shaped by evidence that appears at different positions in a sequence. A review may open with a favorable assessment, develop several positive observations, and reverse its judgment only after a qualifying statement much later in the document. A news article may introduce an entity early and rely on that entity when describing later events. For a compact sequence model, the challenge is therefore not only to encode local expressions, but also to retain information that remains available when subsequent evidence is interpreted. In this study, compact document classification serves as a controlled downstream setting for examining such memory behavior; however, we do not treat the evaluated sequence length as a test of long-context scaling.

Recurrent networks, attention mechanisms, fast-weight systems, and state-space models provide different mechanisms for carrying information through a sequence [1, 4, 6, 11, 13, 15, 37, 41]. When a document classifier jointly optimizes its sequence representation and task classifier, end-task accuracy alone provides limited evidence about the behavior of an intended memory mechanism. The mechanism may never acquire the targeted capability; an acquired capability may be disrupted during task adaptation; the reader may fail to recover stored information; or the classifier may obtain a sufficient decision rule from another pathway. These possibilities require five empirical distinctions: whether a persistent state exists, whether the writer acquires a measurable associative capability, whether that capability survives downstream adaptation, whether a reader can recover sample-specific content, and whether downstream task performance is sensitive to the retrieved content. This distinction is consistent with probing research showing that decodable information need not be functionally important to a model's prediction [36]. Recent causal probing work similarly seeks to distinguish representational encoding from properties that functionally influence model behavior [43]. Representation and internal-intervention methods complement probing by manipulating selected neural state and measuring the resulting change in model behavior [3] [7].

The gap is therefore not simply the absence of another memory mechanism. What remains insufficiently exposed in representative sequence-memory formulations is an architecture and evaluation design that separates these stages empirically. Conventional recurrent encoders provide a bounded state but do not expose distinct writer and reader interfaces. In the cited fast-weight and linear-attention formulations, preserving a previously measured capability during downstream adaptation is not an explicit training objective [1, 15, 37]. In representative memory-augmented formulations, evaluation centers primarily on end-task or reasoning performance rather than on pre-adaptation capability acquisition, post-adaptation retention, sample-specific reader access, and downstream task sensitivity [9, 30, 39, 46]. Without these distinctions, a weak task-level intervention response is difficult to interpret; it may reflect failed capability acquisition, loss of the acquired behavior during adaptation, inaccessible stored content, or genuine task-level insensitivity to an accessible memory state.

We address this gap with Protected QL-Memory (PQLM), a dual-path document classifier that separates local sequence processing from matrix-valued associative state. A causal local-language pathway models short-range token interactions, while a retained matrix writer accumulates sequence-dependent information. After the write phase is complete, a separate reader queries the finalized matrix and combines the retrieved representation with local token state through residual gated fusion. The resulting interfaces expose writing, finalized-state reading, and task prediction to separate measurement and intervention. The term protected is operational. The writer first acquires a controlled binding capability and is then isolated from direct task updates. Because the writer shares the token representation with the local pathway, diagnostic replay constrains that shared interface during local task adaptation. Reader access is subsequently trained through the finalized-memory interface before the task classifier is fitted. The schedule is therefore used as an experimental control: capability acquisition is established before adaptation, preservation is remeasured after adaptation, sample-specific access is tested at the finalized-memory interface, and downstream task sensitivity is evaluated only after these preceding properties have been established. The study addresses four research questions:

- **RQ1:** To what extent does the capability-protected protocol preserve the writer's acquired associative behavior, and does diagnostic access depend on sample-specific finalized memory?
- **RQ2:** How does Protected QL-Memory compare with compact recurrent and attention-based controls in document-classification performance?
- **RQ3:** To what extent is natural-text classification performance sensitive to the selected finalized memory under controlled internal interventions?
- **RQ4:** What parameter, state-storage, runtime, throughput, and accelerator-memory characteristics accompany the protected architecture?

The main contributions of this study are as follows:

1. We formalize memory evaluation as a sequence of distinct empirical claims concerning persistent state, capability acquisition, capability preservation, reader access, and downstream task sensitivity.
2. We propose Protected QL-Memory, a dual-path document classifier with explicit interfaces for causal local processing, recurrent matrix writing, finalized-memory reading, and residual fusion.
3. We introduce a capability-protected training protocol that acquires writer capability before task adaptation, constrains shared-embedding drift through diagnostic replay, and trains finalized-memory access before downstream task fitting, enabling acquisition, preservation, and access failures to be distinguished from task-level insensitivity.
4. We develop an intervention-based evaluation framework that assigns separate measurements to writer retention, reader dependence on sample-specific memory, and natural-text classification-performance sensitivity to the selected finalized state.

We evaluate the proposed framework on three document-classification benchmarks and compare it with compact recurrent and attention-based controls. Classification performance, capability diagnostics, intervention contrasts, and computational characteristics are evaluated separately so that evidence on one dimension is not treated as evidence for another. Section 2 reviews related sequence-memory, associative-memory, and intervention literature. Section 3 formalizes the task and claim structure, Section 4 presents Protected QL-Memory and its capability-protected training protocol, and Section 5 describes the experimental design. Section 6 reports the empirical results, Section 7 discusses their interpretation and limitations, and Section 8 concludes the study.

## 2 RELATED WORK

### 2.1 Compact Sequence Modeling and Persistent State

Recurrent sequence models compress an observed prefix into a state that is updated at each position. Standard recurrent neural networks can be difficult to optimize over long dependency paths [2, 31]. Long Short-Term Memory (LSTM) addresses this problem through a gated cell state with additive updates [13], while the Gated Recurrent Unit (GRU) uses a simpler gated transition without a separate cell state [4]. Large-scale analyses of LSTM variants show that modifying recurrent components can affect performance and identify the forget gate and output activation function as particularly important [10]. These architectures establish bounded-state sequence modeling, but state update and subsequent use remain coupled within the recurrent transition.

Sequence modeling can also be made compact by reducing representation and local-processing costs. Factorized token embeddings replace a direct vocabulary-to-model-width table with a lower-rank vocabulary embedding and projection, reducing the parameter contribution of the embedding table [21]. Local causal operators restrict token interaction to a bounded receptive field, while gated feed-forward transformations such as SwiGLU provide learned multiplicative channel interactions [38]. Recent text-classification architectures continue to combine convolutional feature extraction with recurrent sequence modeling when local and sequential evidence are both relevant [28]. Such components are useful when local linguistic evidence should be represented without requiring the persistent state to perform all token processing.

Transformers replace recurrent state transitions with direct token interactions through self-attention [41]. Standard full attention materializes pairwise scores whose storage grows quadratically with sequence length, motivating sparse and recurrent linear reformulations [15, 18]. Selective state-space models provide another route to recurrent sequence processing. Mamba introduces input-dependent state transitions and a hardware-aware scan [11], while Mamba-2 develops a broader connection between structured state-space models and attention-like computation [6]. These models differ in their state structure and update semantics, illustrating that sequence computation may be organized around vector recurrence, explicit token interaction, or structured recurrent statistics. Memory augmentation has also been combined with modern state-space sequence models to preserve and recover prototype information beyond the recurrent state transition itself [23]. Hierarchical and multi-granularity long-document models address how evidence is aggregated across document levels [24, 25, 40]. The present work focuses on a different distinction: whether a sequence architecture exposes sufficiently separate state-update and read interfaces to measure capability acquisition, preservation, sample-specific access, and downstream task sensitivity independently.

### 2.2 Fast Weights and Explicit Associative Memory

Fast-weight systems introduce states that change within a sequence on a faster timescale than ordinary learned parameters. Ba et al. used rapidly updated weights to attend to the recent hidden state [1]. Schlag et al. subsequently showed that linear Transformers can be interpreted as fast-weight programmers, with outer products writing associations into a matrix-valued state and later queries reading from that state [37]. Linear attention similarly expresses autoregressive attention through recurrent sufficient statistics rather than an explicit token-by-token attention matrix [15]. These formulations motivate matrix-valued recurrence as a mechanism for accumulating sequence-dependent associations. Matrix-valued recurrent states have also been revisited as a means of increasing recurrent storage capacity [14], while associative-network formulations continue to motivate learned update rules for the neural state [27]. Together, these studies reinforce the connection between matrix recurrence and learned associative storage.

These formulations motivate the matrix-valued associative state used in Protected QL-Memory. The relevant distinction for the present study is not matrix storage itself, but the exposure of a finalized recurrent state to a separate post-write reader. Matrix-valued and augmented recurrent designs continue to enlarge the sequence-dependent state beyond a conventional fixed hidden vector [14, 23, 29, 49, 51, 53]. Memory augmentation has also been incorporated into state-space and explicit working-memory architectures [14, 23, 29, 49, 51, 53]. In domain-specific sequential models, dynamic key-value and multi-hop memory structures continue to provide explicit storage and retrieval interfaces [14, 23, 29, 49, 51, 53]. These systems differ substantially in update semantics and task setting, but collectively demonstrate continued interest in separating sequence processing from dedicated memory state. A related line of work provides differentiable external memory with explicit learned storage and retrieval interfaces. Memory Networks combine learned inference components with addressable memory [46], and End-to-End Memory Networks train multi-hop access from input-output supervision [9, 30, 39, 46]. Key-Value Memory Networks separate addressing keys from stored values for document reading [30]. Differentiable Neural Computers couple neural controllers to external memory through learned read and write operations, allocation, and temporal linkage [9]. These systems demonstrate the value of explicit memory interfaces, but their addressable external-memory organization and task settings differ from the fixed ($r \times r$) recurrent matrix examined here.

Associative-memory interpretations also connect attention with modern Hopfield networks [35]. This connection provides an account of storage and retrieval in energy-based systems. Protected QL-Memory instead performs one causal write scan followed by a post-write finalized-memory read phase. It does not retain all token state for softmax attention and does not perform iterative Hopfield retrieval.

### 2.3 Probing, Intervention, and Capability Preservation

Representation probing asks whether selected information can be recovered from an internal state. Controlled studies have shown that high probe accuracy does not necessarily imply that the task model relies on the decoded property [36]. Amnesic probing and related intervention methods therefore, examine how model behavior changes when internal information is removed or altered [7]. This distinction motivates separate writer diagnostics, reader interventions, and natural-text memory interventions in the present study. A related concern is interference between an acquired capability and subsequent optimization. Forgetting is increasingly treated as a broader deep-learning phenomenon in which previously acquired knowledge or behavior may deteriorate under later learning, rather than as a problem restricted to canonical continual-learning benchmarks [45]. Replay, regularization, and architectural or parameter-isolation strategies are established approaches for limiting such interference [16, 17, 42, 44, 50] and studies of continual pre-training further show that retention can depend on the subsequent adaptation process and training protocol [5]. The preservation problem considered here is narrower: one measured writer capability is acquired before a shared token representation adapts to document classification. Protected QL-Memory therefore combines writer isolation with diagnostic replay and directly measures the target behavior before and after adaptation. Accordingly, diagnostic replay is not presented as a new continual-learning method; it is used with writer isolation as a preservation control before testing downstream memory sensitivity.

The combination of probing and preservation raises a further distinction. A capability may remain intact under the controlled diagnostic after adaptation, while downstream classification performance remains insensitive to the selected finalized state. The proposed evaluation, therefore, treats capability preservation, reader access, and downstream task sensitivity as separate empirical properties rather than successive names for the same outcome.

### 2.4 Positioning of Protected QL-Memory

Recurrent, attention-based, and state-space encoders provide different forms of state transition or token interaction [4, 6, 11, 13, 41], while fast-weight and memory-augmented models expose mechanisms for writing and retrieving sequence-dependent information [1, 9, 15, 30, 37, 39, 46]. A separate probing and intervention literature has shown that recoverable information need not be functionally important to model behavior and has increasingly used internal interventions to test behavioral sensitivity to represented information [3, 7, 33, 36, 43]. Protected QL-Memory connects these lines through an explicit writer-reader interface, a staged capability-preservation protocol, and a post-write intervention point. The purpose of this combination is diagnostic: a weak downstream intervention response is interpreted only after capability acquisition, preservation, and sample-specific access have been examined separately.

Table 1. Positioning relative to representative formulations of major sequence-memory families. Entries summarize the cited formulations rather than every architecture within each family

| Model family | Typical sequence state | Read/write structure | Explicit capability-preservation objective | Intervention interface |
|---|---|---|---|---|
| LSTM/GRU [4, 13] | Fixed hidden or cell vector | Coupled recurrent transition | Not a defining objective of the cited formulations | No separate post-write interface |
| Fast weights / linear attention [1, 15, 37] | Recurrent statistics or matrix | Associative update and query read | Not a defining objective of the cited formulations | Formulation dependent |
| External-memory models [9, 30, 39, 46] | Addressable memory with selected capacity | Explicit learned read/write operations | Not a defining objective; formulation dependent | Architecture dependent |
| Protected QL-Memory | Fixed recurrent matrix | Separate the writer and the finalized-state reader | Writer isolation + diagnostic replay; pre/post retention measured | Post-write, pre-read interface |

The contribution of Protected QL-Memory is therefore not a new claim that recurrent matrices can store associations. The contribution is likewise not a new replay rule or checkpointing strategy. Rather, writer isolation and diagnostic replay are used as preservation controls within a framework that connects an explicit write-read interface with separate measurements of acquired capability, post-adaptation retention, sample-specific reader access, and downstream task sensitivity under finalized-memory intervention. This positioning shifts the primary question from whether information can exist in an internal state to whether a measured capability survives adaptation, remains accessible through the finalized-memory interface, and produces an observable change in downstream classification performance when the selected state is altered.

## 3 PROBLEM FORMULATION AND CLAIM STRUCTURE

This section formalizes the document-classification task and the empirical claims used to evaluate the memory mechanism. We first define the task notation, then introduce the capability and intervention contrasts, and finally map each claim to the evidence required to support it.

### 3.1 Document Classification Task

For document $i$, let $x^{(i)}$ denote a sequence of token identifiers from a vocabulary of size $V$. Let $m^{(i)} \in \{0,1\}^T$ denote a right-padded prefix mask, where $m_t^{(i)} = 1$ identifies a valid token position and $m_t^{(i)} = 0$ identifies padding. Let $y^{(i)} \in \{1, \dots, C\}$ denote the document label. The token sequence and number of valid positions are represented as

$$x^{(i)} = \left(x_1^{(i)}, \dots, x_T^{(i)}\right), x_t^{(i)} \in \{0, \dots, V-1\}, \qquad (1)$$

$$n^{(i)} = \sum_{t=1}^{T} m_t^{(i)}. \qquad (2)$$

For notational simplicity, the document superscript $(i)$ is omitted below when referring to a generic example. Given x and m, an encoder-classifier parameterized by θ produces class logits $z_\theta(x,m) \in \mathbb{R}^C$. The corresponding class probabilities and predicted label are as defined in Eq. (3). The supervised document-classification objective is multiclass cross-entropy over the training set $\mathcal{D}_{\text{tr}}$ as defined by Eq. (4). Table 2 summarizes the principal notation used in the architecture and evaluation.

$$p_\theta(y = c \mid x, m) = \frac{\exp\left(z_{\theta,c}(x,m)\right)}{\sum_{j=1}^{C} \exp\left(z_{\theta,j}(x,m)\right)}, \hat{y} = \arg\max_{1 \le c \le C} p_\theta(y = c \mid x, m). \qquad (3)$$

$$\mathcal{L}_{\text{task}}(\theta) = -\frac{1}{\mid \mathcal{D}_{\text{tr}} \mid} \sum_{(x,m,y) \in \mathcal{D}_{\text{tr}}} \log p_\theta(y \mid x, m). \qquad (4)$$

*Table 2. Principal notation*

| Symbol | Meaning |
|---|---|
| $x, m, y, n$ | Token sequence, prefix mask, document label, and valid-token count |
| $V, T, C$ | Vocabulary size, maximum input length, and number of classes |
| $\theta, z_\theta, p_\theta, \hat{y}$ | Model parameters, class logits, class probabilities, and predicted label |
| $\mathcal{D}_{\text{tr}}$ | Document-classification training set |
| $r_e, d, r, k$ | Embedding factor rank, model width, matrix width, and local kernel width |
| $e_t, u_t, \ell_t$ | Token representation, causal pair context, and local-language state |
| $M_t, M_T, \widetilde{M}_T$ | Recurrent matrix, finalized matrix, and selected or intervened finalized matrix |
| $h_t^g$ | Writer output state after the internal matrix read |
| $a_t, w_t$ | Association factor and row-write factor |
| $q_t^g, q_t^r$ | Writer-internal query and finalized-memory query |
| $\lambda_t, \beta_t$ | Row-wise retention and write gate |
| $\rho_t, r_t$ | Raw matrix readout and projected finalized-memory state |
| $\gamma_t, c_t$ | Fusion gate and memory-conditioned token state |
| $A_{\text{pre}}, A_{\text{post}}$ | Writer diagnostic accuracy before and after local-language adaptation |
| $\Delta A_{\text{cap}}, R_{\text{cap}}$ | Absolute capability change and post/pre capability-retention ratio |
| $A_{\text{reader}}^{\text{normal}}, A_{\text{reader}}^{\text{shift}}$ | Reader diagnostic accuracy under normal and cyclically shifted finalized memory |
| $U_{\text{reader}}$ | Normal-minus-shift reader diagnostic contrast |
| $\mu, F_\mu$ | Memory condition and validation macro-F1 under that condition |
| $U_{\text{task}}^{\text{shift}}, U_{\text{task}}^{\text{zero}}, U_{\text{task}}^{\text{mean}}$ | Normal-minus-intervention macro-F1 contrasts |
| $F_{\text{local}}, \Delta_{\text{route}}$ | Selected local-route macro-F1 and fused-minus-local route-performance contrast |

### 3.2 Memory Claims and Intervention Contrasts

The presence of a recurrent matrix establishes that the architecture contains a persistent within-document state. Here, persistent refers to a state carried across token positions within a document; the matrix is not shared across documents. This structural property does not, however, determine what information the state acquires, whether an acquired behavior survives subsequent optimization, whether diagnostic access is sensitive to example-memory correspondence, or whether aggregate document-classification performance is sensitive to the selected state under intervention. Motivated by the distinction between representational recoverability and functional relevance in probing and intervention studies [7, 36], we treat these as separate empirical claims. We therefore distinguish four learned properties: writer acquisition, capability preservation, diagnostic access, and downstream task-performance sensitivity. These properties are evaluated separately so that evidence for one is not used as a proxy for another.

**Writer acquisition and capability preservation**

Writer acquisition is measured on the controlled binding task immediately after writer training and before local-language adaptation. Let $A_{\text{pre}}$ denote this diagnostic accuracy. After the shared embedding and local-language pathway have adapted to document classification, the same writer diagnostic is evaluated again, yielding $A_{\text{post}}$. Capability preservation is summarized by the absolute diagnostic change and the relative post/pre ratio:

$$\Delta A_{\text{cap}} = A_{\text{post}} - A_{\text{pre}}, R_{\text{cap}} = \frac{A_{\text{post}}}{A_{\text{pre}}}. \qquad (5)$$

The absolute difference records the direction and magnitude of diagnostic change, whereas the ratio expresses post-adaptation performance relative to the pre-adaptation score. Values of $R_{\text{cap}} > 1$ indicate that post-adaptation diagnostic accuracy exceeded the initial score; they are not interpreted as retention beyond 100%. Because the shared embedding and writer interface are fixed after Stage 1, subsequent training stages do not modify the components evaluated by the writer diagnostic.

**Diagnostic access to sample-specific finalized memory**

The finalized-memory pathway is evaluated through the controlled diagnostic described in Section 4.7. Importantly, the reader diagnostic head does not operate directly on the raw matrix read. It receives the terminal fused representation after finalized-memory querying, memory projection, gated residual fusion, and full-path normalization.

Let $A_{\text{reader}}^{\text{normal}}$ denote diagnostic accuracy when each example uses its own finalized matrix. Let $A_{\text{reader}}^{\text{shift}}$ denote the corresponding accuracy when finalized matrices are cyclically reassigned across examples while the reader and diagnostic head remain fixed. The reader intervention contrast is

$$U_{\text{reader}} = A_{\text{reader}}^{\text{normal}} - A_{\text{reader}}^{\text{shift}}. \qquad (6)$$

A positive $U_{\text{reader}}$ records higher diagnostic accuracy when the example-memory correspondence is preserved than when it is broken by cyclic reassignment. The magnitude of the contrast quantifies sensitivity to that correspondence; the predefined acceptance threshold is specified in Section 5.4. This contrast concerns diagnostic access through the terminal fused representation and does not require the controlled target to be directly decodable from the raw matrix read $\rho_t$ alone.

**Downstream task-performance sensitivity**
Task-performance sensitivity is evaluated separately on natural-text validation data. The intervention is applied after the writer scans and before finalized-memory reading, while all post-intervention parameters and operations are held fixed.
Let $\mu \in \{\text{normal}, \text{shift}, \text{zero}, \text{mean}\}$ denote the evaluated memory condition, and let $F_\mu$ denote validation macro-F1 under condition $\mu$. The primary cyclic-reassignment contrast is

$$U_{\text{task}}^{\text{shift}} = F_{\text{normal}} - F_{\text{shift}}. \qquad (7)$$

Two complementary contrasts replace the finalized matrix with zero or the batch mean:

$$U_{\text{task}}^{\text{zero}} = F_{\text{normal}} - F_{\text{zero}}, U_{\text{task}}^{\text{mean}} = F_{\text{normal}} - F_{\text{mean}}. \qquad (8)$$

Under this sign convention, positive values indicate lower macro-F1 under intervention than under normal memory, whereas negative values indicate higher macro-F1 under intervention. The absolute magnitude records the observed performance sensitivity. Equations (6) to (8) define within-model intervention contrasts. They quantify the sensitivity of the specified diagnostic or aggregate classification metric to controlled replacement of the finalized state, while the remaining computation is held fixed [3]. They do not estimate causal effects in the data-generating process.

**Preservation of the local task route**
Let $F_{\text{local}}$ denote the validation macro-F1 of the selected local-route checkpoint and let $F_{\text{normal}}$ denote validation macro-F1 of the selected full-path checkpoint under normal memory. Their fused-minus-local route-performance contrast is

$$\Delta_{\text{route}} = F_{\text{normal}} - F_{\text{local}}. \qquad (9)$$

A protocol-specific tolerance for $\Delta_{\text{route}}$ is defined in Section 5.4. Because the local and full-path checkpoints are selected under their respective validation objectives, $\Delta_{\text{route}}$ is a protocol-level route-performance contrast rather than an internal memory intervention. It records whether the selected fused pathway preserves performance relative to the selected local route. A fused model may satisfy this criterion while remaining insensitive to changes in the finalized matrix.

### 3.3 Claim-to-Evidence Mapping

The measurements above assign different evidential requirements to different claims. Table 3 summarizes this mapping.

Table 3. Claim-to-evidence mapping

| Empirical question | Primary evidence | Interpretive scope |
|---|---|---|
| Does persistent within-document state exist? | Architecture definition and mask-safe state tests | Structural property |
| Did the writer acquire controlled binding? | $A_{\text{pre}}$ | Controlled writer capability |
| Was the measured capability preserved? | $A_{\text{post}}, \Delta A_{\text{cap}}, R_{\text{cap}}$ | Pre/post behavior on the controlled writer diagnostic |
| Is diagnostic performance sensitive to example-memory correspondence? | $U_{\text{reader}}$ from normal versus cyclic-shift finalized memory | Sensitivity of the terminal fused diagnostic representation to sample-specific finalized state |
| Is aggregate classification performance sensitive to the selected finalized state? | $U_{\text{task}}^{\text{shift}}, U_{\text{task}}^{\text{zero}}, U_{\text{task}}^{\text{mean}}$ | Within-model macro-F1 sensitivity under specified finalized-memory interventions |
| Does the selected full-path route preserve local-route performance within protocol tolerance? | $\Delta_{\text{route}} = F_{\text{normal}} - F_{\text{local}}$ | Protocol-level route-performance contrast |

The mapping prevents evidential substitution across claims. The architecture establishes the existence and dimensions of the recurrent state. Writer diagnostic accuracy measures acquisition of the controlled binding behavior, and the pre/post comparison measures preservation of that behavior during local-language adaptation. The diagnostic memory intervention evaluates whether performance through the terminal fused diagnostic representation is sensitive to example-memory correspondence. Natural-text interventions separately evaluate whether validation macro-F1 is sensitive to the replacement of the selected finalized state. No single result substitutes for the others. In particular, strong writer accuracy does not establish reader access, and strong diagnostic access does not establish downstream task-performance sensitivity. The claims are restricted to compact document classification under the evaluated input length, architecture, and staged training protocol. The controlled binding task measures a specific acquired writer behavior and diagnostic access through the finalized-memory fused representation. It is not intended as a general measure of semantic memory or long-context reasoning. Natural-text task sensitivity is evaluated separately through interventions on the selected finalized matrix. Throughout this study, downstream task utilization is operationalized as measurable sensitivity of validation macro-F1 to finalized-memory intervention. This metric-bounded definition does not imply that individual predicted labels or class-probability vectors are invariant when macro-F1 is unchanged. Accordingly, diagnostic success is interpreted as evidence for the measured capability or access property, whereas the downstream task claim is restricted to the observed macro-F1 intervention contrasts.

# 4 PROTECTED QL-MEMORY

This section specifies the Protected QL-Memory forward architecture, the post-write intervention interface, and the capability-protected training schedule. We first define the local and matrix pathways, then formalize finalized-memory reading and evaluation-time intervention and finally describe stage-restricted optimization and the computational characteristics of the architecture.

## 4.1 Architecture Overview and Design Principles

Protected QL-Memory maps a masked token sequence through a shared factorized token representation and two distinct pathways, as illustrated in Fig. 1. The local-language pathway produces token states $\ell_t \in \mathbb{R}^d$. The matrix pathway transforms a causal pair context into writer states $h_t^g \in \mathbb{R}^d$and a finalized recurrent matrix $M_T \in \mathbb{R}^{r\times r}$. After the write scan is complete, the normal forward path uses the document's own finalized matrix $M_T$. During intervention-based evaluation, an evaluation-only operator may instead select an alternative finalized matrix $\widetilde{M}_T$. Each valid local state then generates a query for the same selected matrix. The resulting readout is projected to the model width, fused residually with the local state, and pooled to form the document representation.

The architecture follows four design principles. First, the factorized embedding, pair-context projection, local pathway, matrix writer, finalized-memory reader, fusion gate, and classifier are organized into separable parameter groups that can be optimized or held fixed at different stages. Second, the pair context and local convolution are causal with respect to token access, and masked positions cannot modify the recurrent matrix. Third, the pair-context projection and matrix writer are frozen after capability acquisition, while diagnostic replay constrains adaptation of the shared embedding through which the fixed writer interface receives token information. Fourth, the finalized matrix is exposed at a stable post-write, pre-read interface, allowing its content to be replaced while all post-intervention parameters and operations remain fixed.

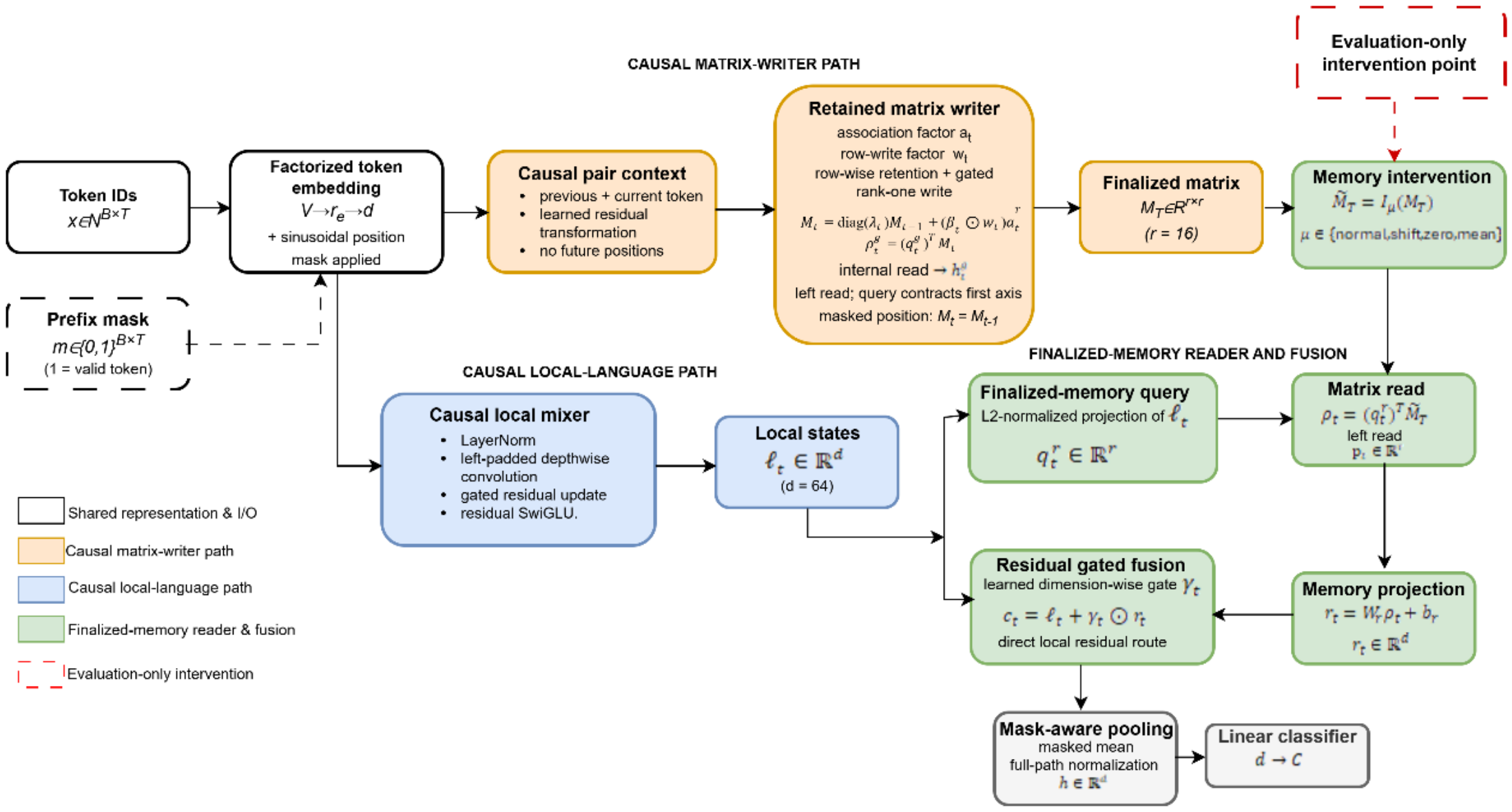

**Fig.** 1 Protected QL-Memory forward architecture**.** A shared factorized token representation feeds a causal local-language pathway and a retained matrix writer. The writer performs row-wise retention and gated rank-one updates to the recurrent matrix $M_t$. After the write scan, the normal path uses $M_T$; during intervention-based evaluation, an evaluation-only operator may replace it with a selected finalized matrix $\widetilde{M}_T$before finalized-memory reading. Local states generate queries for the selected matrix and also provide a direct residual route to gated fusion. Both writer-internal and finalized-memory reads contract the first matrix axis

## 4.2 Factorized Embedding and Causal Pair Context

Following the factorized embedding principle used to separate vocabulary embedding width from model hidden width [21], let $A \in \mathbb{R}^{V\times r_e}$ be a learned code table and $W_e \in \mathbb{R}^{d\times r_e}$ a bias-free projection. With sinusoidal position vector $p_t$ [41], the masked token representation is

$$e_t = m_t(W_e A[x_t] + p_t) \in \mathbb{R}^d. \qquad (10)$$

Because the mask multiplies the complete token-plus-position representation, $e_t = 0$ at padded positions. The writer also requires a causal transition-level signal that distinguishes the current token from its immediate predecessor without exposing future positions. We therefore augment the current token representation with a learned transformation of the previous-current token pair:

$$u_t = m_t\{e_t + \mathrm{SiLU}(W_c[e_{t-1}; e_t] + b_c)\}, e_0 = 0. \qquad (11)$$

Because $u_t$ depends only on $e_{t-1}$and $e_t$, the pair context contains no token representation from positions after $t$. The outer mask also gives $u_t = 0$ at padded positions.

### 4.3 Causal Local-Language Pathway

The local pathway captures bounded-range token interactions independently of the recurrent matrix. The token sequence is first normalized and processed by a depthwise convolution after $k - 1$zero vectors are added to the left:

$$s_{1:T} = \mathrm{DWConv}_k\big(\mathrm{LeftPad}_{k-1}(\mathrm{LN}(e_{1:T}))\big). \qquad (12)$$

The left padding ensures that $s_t$depends only on positions $t - k + 1$through $t$. A pointwise projection produces a content branch and gate logits, followed by a residual update:

$$[a_t^{\mathrm{loc}}; b_t^{\mathrm{loc}}] = W_{cg} s_t + b_{cg},$$
$$v_t = e_t + m_t \mathrm{Drop}\big(W_{co}\big[\mathrm{SiLU}(a_t^{\mathrm{loc}}) \odot \sigma(b_t^{\mathrm{loc}})\big] + b_{co}\big). \qquad (13)$$

A second residual block applies a SwiGLU transformation [38] to refine the channel representation:

$$[f_t; g_t] = W_{ff}\mathrm{LN}(v_t) + b_{ff},$$
$$\ell_t = v_t + m_t \mathrm{Drop}\big(W_{fo}[f_t \odot \mathrm{SiLU}(g_t)] + b_{fo}\big). \qquad (14)$$

Here, Dropdenotes dropout and $\odot$ denotes element-wise multiplication. Because $e_t = 0$ at padding and both residual updates are mask-gated, padded positions remain zero in the local representation. The resulting state $\ell_t \in \mathbb{R}^d$ forms the local route used for Stage 1 document classification and later serves as the query source and residual input to finalized-memory fusion.

### 4.4 Retained Matrix Writer

Motivated by associative fast-weight formulations in which sequence-dependent information is accumulated through matrix-valued state and outer-product updates [1, 37], the proposed writer maintains a recurrent matrix $M_t \in \mathbb{R}^{r\times r}$, initialized by $M_0 = 0$. The implemented writer combines normalized association factors, bounded row-write content, learned row-wise retention, and gated rank-one updates within the protected writer interface. From the causal pair context, the writer first computes a shared expanded feature:

$$\phi_t = \mathrm{SiLU}(W_s \mathrm{LN}(u_t) + b_s). \qquad (15)$$

The branch split follows the implemented matrix axes. It produces an association-factor precursor, a row-write precursor, a writer-internal query precursor, row-retention logits, write-gate logits, and output-gate logits:

$$[\bar{a}_t; \bar{w}_t; \bar{q}_t^g; a_t^\lambda; a_t^\beta; o_t] = W_b \phi_t + b_b. \qquad (16)$$

For numerically safe $\ell_2$ normalization, define

$$\mathrm{Norm}_\varepsilon(v) = \frac{v}{\max(\| v \|_2, \varepsilon)}, \qquad (17)$$

where $\varepsilon > 0$ is the fixed numerical-stability constant used by the normalization operation. The writer signals and their effective mask-safe forms are

$$a_t = \mathrm{Norm}_\varepsilon(\bar{a}_t), w_t = \tanh(\bar{w}_t), q_t^g = \mathrm{Norm}_\varepsilon(\bar{q}_t^g),$$
$$\lambda_t = 0.90 + 0.099\sigma(a_t^\lambda), \beta_t = \sigma(a_t^\beta),$$
$$\lambda_t^{\mathrm{eff}} = (1 - m_t)\mathbf{1} + m_t\lambda_t, \omega_t^{\mathrm{eff}} = m_t(\beta_t \odot w_t), q_t^{g,\mathrm{eff}} = m_t q_t^g. \qquad (18)$$

The affine sigmoid parameterization restricts each unmasked row-retention coefficient to [0.900, 0.999]. It therefore limits abrupt per-step erasure of a matrix row while retaining a learned decay rate within the range used by the evaluated architecture. The write gate $\beta_t$ controls the dimension-wise magnitude of the new row contribution. The recurrent matrix applies effective row-wise retention followed by a gated rank-one update:

$$M_t = \mathrm{diag}\big(\lambda_t^{\mathrm{eff}}\big)M_{t-1} + \omega_t^{\mathrm{eff}} a_t^\top. \qquad (19)$$

Here, $\omega_t^{\mathrm{eff}}$ occupies the first matrix axis and $a_t$occupies the second matrix axis. At a masked position, $m_t = 0$, so $\lambda_t^{\mathrm{eff}} = 1$, $\omega_t^{\mathrm{eff}} = 0$, and $q_t^{g,\mathrm{eff}} = 0$. Equation (19), therefore gives $M_t = M_{t-1}$. Padding cannot modify the persistent within-document matrix state. The writer-internal query reads the updated matrix from the left:

$$\rho_t^g = (q_t^{g,\text{eff}})^\top M_t,$$
$$h_t^g = u_t + m_t \text{Drop}\{(W_o^g \rho_t^g + b_o^g) \odot \sigma(o_t)\}. \quad (20)$$

Both matrix readers contract the first matrix axis. For a single rank-one write,

$$q^\top[(\beta \odot w)a^\top] = [q^\top(\beta \odot w)]a^\top. \quad (21)$$

Thus, left contraction of the first matrix axis produces a scalar coefficient multiplying $a^\top$, and the raw read lies in the association-factor direction. The terms row-write factor and association factor describe these implemented axes without implying the conventional key-to-value orientation $M = vk^\top$ followed by $Mq$. The output gate controls, dimension by dimension, how strongly the internal matrix read modifies the causal pair representation. During writer acquisition, the writer state $h_t^g$ at the terminal diagnostic query position is supplied to the writer diagnostic head, as described in Section 4.7.

### 4.5 Finalized-Memory Reader and Fusion

Query-based access to stored state is a recurring principle in associative fast-weight and explicit key-value memory systems [30, 37]. Protected QL-Memory applies this principle at a distinct post-write interface. The recurrent writer first completes its scan, after which each local-language state generates a normalized query for the selected finalized matrix:

$$q_t^r = \text{Norm}_\varepsilon(W_q^r \ell_t + b_q^r). \quad (22)$$

The finalized-memory query $q_t^r$ is distinct from the writer-internal query $q_t^g$. It contracts the same first matrix axis and projects the resulting second-axis readout to the model width:

$$\rho_t = (q_t^r)^\top \widetilde{M}_T, r_t = W_r \rho_t + b_r. \quad (23)$$

The projected readout is integrated with the local representation through dimension-wise gated residual fusion. The fusion gate is conditioned jointly on the local and retrieved states:

$$\gamma_t = \sigma(W_\gamma[\ell_t; r_t] + b_\gamma), c_t = \ell_t + \gamma_t \odot r_t. \quad (24)$$

The residual term provides a direct local route, while the gated term allows the finalized-memory pathway to modify individual representation dimensions. The additive bypass retains the local state as an explicit residual component, but it also permits a classification solution that is insensitive to the selected finalized matrix. Downstream task-performance sensitivity to that matrix is therefore evaluated explicitly through the interventions defined in Section 4.6. All valid positions query the same selected finalized matrix. For a document with $n \geq 1$valid tokens, masked mean pooling and full-path normalization produce the document representation and class logits:

$$h = \text{LN}_{\text{full}}\left(\frac{1}{n}\sum_{t=1}^{T} m_t\, c_t\right), z = W_{\text{cls}} h + b_{\text{cls}}. \quad (25)$$

Although intermediate values may be computed at padded positions, the masked pooling operation excludes them from the document representation. The writer and local convolution are causal with respect to sequence access. Document classification, however, occurs only after the write phase has completed and the finalized matrix is available to the reader.

### 4.6 Evaluation-Time Finalized-Memory Intervention

Representation and internal-intervention methods evaluate model sensitivity by manipulating selected internal states while holding the remaining computation fixed [3, 7, 34]. The separation between writing and finalized-memory reading exposes a stable post-write, pre-read intervention point in Protected QL-Memory. For a batch of finalized matrices $\{M_T^{(i)}\}_{i=1}^{B}$, an intervention condition $\mu \in \{\text{normal}, \text{shift}, \text{zero}, \text{mean}\}$ selects

$$\widetilde{M}_T^{(i)} = \mathcal{I}_\mu(M_T^{(i)}) = \begin{cases} M_T^{(i)}, & \mu = \text{normal}, \\ M_T^{1+((i-2) mod B)}, & \mu = \text{shift}, \\ 0, & \mu = \text{zero}, \\ \frac{1}{B}\sum_{j=1}^{B} M_T^{(j)}, & \mu = \text{mean}. \end{cases} \quad (26)$$

For the cyclic-shift condition, $B \geq 2$ is required. The shift condition deterministically assigns each example the preceding example's finalized matrix under cyclic indexing. This cyclic reassignment preserves the dimensional structure and scale characteristics of an empirically produced finalized state while breaking the original example-memory correspondence. The zero condition removes the finalized-matrix contribution. The batch-mean condition replaces sample-specific state with shared batch-level content. The intervention is applied after the writer scans and before any finalized-memory query. All post-intervention parameters and operations are held fixed across memory conditions. The resulting contrasts, therefore, quantify the sensitivity of the specified diagnostic or aggregate classification metric to the selected finalized state. Section 5 specifies the primary and complementary roles of these conditions in the empirical protocol.

### 4.7 Capability-Protected Training

Protected QL-Memory is optimized in four restricted stages, illustrated in Fig. 2. The schedule is used as an experimental control rather than as a general continual-learning procedure. Its ordering is deliberate: writer capability is established before downstream adaptation, preservation is measured after shared-representation adaptation, diagnostic access is trained through the finalized-memory interface, and document-label fitting remains restricted until these preceding properties have been established.

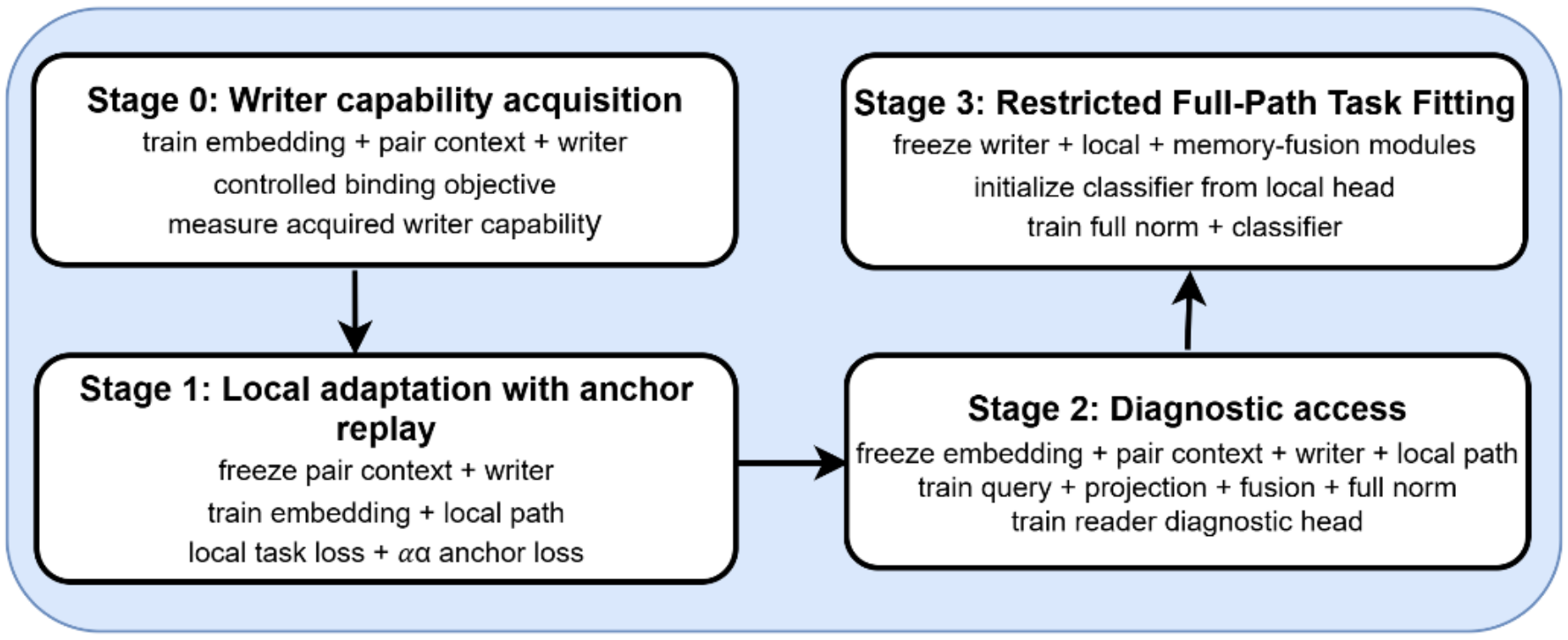

**Fig.** 2 Capability-protected training flow. Writer capability is acquired before local-language adaptation. Diagnostic replay constrains changes to the shared embedding during local adaptation. Diagnostic access is then trained through the finalized-memory fused representation, and the task classifier is fitted only after the writer, local pathway, and memory-reading and fusion modules are fixed. Each stage restricts optimization to a designated parameter group

**Stage 0: Writer Capability Acquisition**

Motivated by associative key-value retrieval formulations in fast-weight and explicit memory systems [30, 37]; the Stage 0 acquires a controlled binding capability before document supervision is introduced. The diagnostic construction is study-specific. A sequence contains key-value token pairs, corpus-derived distractor tokens, and a terminal query that identifies one previously presented key; the target is the associated value. Here, *key* and *value* describe the controlled input-target relation rather than the internal axes of the recurrent matrix. The factorized embedding, causal pair-context projection, matrix writer, and writer diagnostic head are optimized on the controlled task. Let $t^\star$ denote the final valid position, which contains the terminal key query. The writer diagnostic head receives $h^{g}_{t^\star}$ and predicts the associated value through categorical cross-entropy. Stage 0 consequently produces a writer whose acquired binding behavior can be measured before downstream adaptation.

**Stage 1: Local-Language Adaptation with Diagnostic Replay**

Replay and parameter-isolation strategies are established approaches for limiting interference with previously acquired behavior during subsequent learning [42, 44, 50]. The preservation problem considered here is narrower: Protected QL-Memory seeks to retain one explicitly measured writer capability while a shared token representation adapts to document classification. The pair-context projection and matrix writer are frozen, while the factorized embedding, local pathway, local-route normalization, and local auxiliary classifier remain trainable. Freezing the writer alone does not fully isolate the acquired behavior because the embedding is shared with the local pathway. Changes to that representation can alter the effective writer input distribution even when the writer parameters are unchanged. For the local task route, the masked local states are pooled and normalized as

$$h_{\text{local}} = \mathrm{LN}_{\text{local}}\left(\frac{1}{n}\sum_{t=1}^{T} m_t\, \ell_t\right), \qquad (27)$$

with local-route logits

$$z_{\text{local}} = W_{\text{local}} h_{\text{local}} + b_{\text{local}}. \quad (28)$$

Each document-task batch is accompanied by a controlled diagnostic replay batch. Stage 1 minimizes

$$\mathcal{L}_1 = \mathcal{L}_{\text{local-task}} + \alpha \mathcal{L}_{\text{anchor}}. \qquad (29)$$

Here, $\mathcal{L}_{\text{local-task}}$ is document-label cross-entropy from the local auxiliary classifier, and $\mathcal{L}_{\text{anchor}}$ is categorical cross-entropy from the fixed writer diagnostic head on a replayed binding batch. Because the pair-context projection, matrix writer, and writer diagnostic head are fixed, the gradient of $\mathcal{L}_{\text{anchor}}$reaches trainable parameters through the shared factorized embedding. The replay term, therefore, constrains embedding changes that disrupt the acquired writer behavior. Writer diagnostic accuracy is measured immediately before and after Stage 1, providing $A_{\text{pre}}$ and $A_{\text{post}}$ for the capability-preservation contrasts in Eq. (5). Because the shared embedding and writer interface are fixed after Stage 1, subsequent training stages do not modify the components evaluated by the writer diagnostic.

**Stage 2: Diagnostic Access through Finalized-Memory Fusion**

After local adaptation, the factorized embedding, pair-context projection, matrix writer, local pathway, and local-route normalization are fixed. Stage 2 trains only the finalized-memory query, memory-output projection, fusion gate, full-path normalization, and reader diagnostic head. For each controlled binding sequence, all valid local states query the same finalized matrix. Let $t_i^{\star} = n^{(i)}$ denote the final valid position of the example $i$, corresponding to the terminal queried-key position in the diagnostic sequence. The reader diagnostic head receives the full-path-normalized terminal fused state

$$h_i^{r,\text{diag}} = \text{LN}_{\text{full}}\left(c_{i,t_i^{\star}}\right). \quad (30)$$

Its logits and class probabilities are

$$z_i^{r,\text{diag}} = W_{r,\text{diag}} h_i^{r,\text{diag}} + b_{r,\text{diag}}, p_{\text{reader},i} = \text{softmax}\left(z_i^{r,\text{diag}}\right). \quad (31)$$

For a controlled diagnostic set $\mathcal{D}_{\text{diag}}$ with value target $y_a^{(i)}$, Stage 2 minimizes

$$\mathcal{L}_2 = -\frac{1}{|\mathcal{D}_{\text{diag}}|} \sum_{(x^{(i)}, m^{(i)}, y_a^{(i)}) \in \mathcal{D}_{\text{diag}}} \log p_{\text{reader},i}\left[y_a^{(i)}\right]. \quad (32)$$

Equation (32) supervises the terminal fused representation after finalized-memory reading, memory projection, gated residual fusion, and full-path normalization. Stage 2, therefore, trains controlled diagnostic access through the finalized-memory pathway. Normal reader accuracy at the end of Stage 2 is retained as a training-progress diagnostic. It is not the final reader-access contrast used for the protected-run claim. Because Stage 3 subsequently updates full-path normalization, $A_{\text{reader}}^{\text{normal}}$, $A_{\text{reader}}^{\text{shift}}$, and $U_{\text{reader}}$ are recomputed after the selected Stage 3 checkpoint is reloaded. The final $U_{\text{reader}}$ therefore characterizes sensitivity to example-memory correspondence at the selected downstream checkpoint rather than a pre-Stage-3 progression gate.

**Stage 3: Restricted Full-Path Task Fitting**

The main classifier is initialized from the local auxiliary classifier. All modules except full-path normalization and the main classifier are then frozen. Only these two components are fitted to document labels. This restriction prevents Stage 3 document-label supervision from modifying the acquired writer interface, the local representation pathway, or the learned finalized-memory reading and fusion modules. Full-path normalization remains trainable together with the final decision boundary. The selected full-path checkpoint is subsequently evaluated under the capability, reader-access, and route-performance criteria defined in Section 5. Algorithm 1 presents the steps for the capability-protected training of Protected QL-Memory, and Table 4 shows the module-specific optimization status across each training schedule.

Table 4. Module-specific optimization status across the capability-protected training schedule

| **Module** | **Stage 0** | **Stage 1** | **Stage 2** | **Stage 3** |
|---|---|---|---|---|
| Factorized embedding | Train | Train | Frozen | Frozen |
| Pair-context projection | Train | Frozen | Frozen | Frozen |
| Matrix writer | Train | Frozen | Frozen | Frozen |
| Local pathway and local-route normalization | Frozen | Train | Frozen | Frozen |
| Finalized-memory query and memory-output projection | Frozen | Frozen | Train | Frozen |
| Fusion gate | Frozen | Frozen | Train | Frozen |
| Full-path normalization | Frozen | Frozen | Train | Train |
| Main classifier | Frozen | Frozen | Frozen | Train |
| Local auxiliary classifier | Frozen | Train | Frozen | Frozen |
| Writer diagnostic head | Train | Frozen | Frozen | Frozen |
| Reader diagnostic head | Frozen | Frozen | Train | Frozen |

**Algorithm 1. Capability-protected training of Protected QL-Memory**

**Input:** Task data $\mathcal{D}$, controlled diagnostic generator $\mathcal{G}_{\text{diag}}$, diagnostic-replay loss weight $\alpha$

**Output:** Selected Protected QL-Memory model and protocol diagnostics

1. Train the factorized embedding, pair-context projection, matrix writer, and writer diagnostic head on samples from $\mathcal{G}_{\text{diag}}$.
2. Measure the acquired writer capability $A_{\text{pre}}$.
3. Freeze the pair-context projection and matrix writer.
4. Adapt the factorized embedding, local pathway, local-route normalization, and local auxiliary classifier using $\mathcal{L}_1 = \mathcal{L}_{\text{local-task}} + \alpha\mathcal{L}_{\text{anchor}}$.
5. Re-evaluate writer capability and record $A_{\text{post}}$, $\Delta A_{\text{cap}}$, and $R_{\text{cap}}$.
6. Freeze the factorized embedding, local pathway, and local-route normalization.
7. Train the finalized-memory query, memory-output projection, fusion gate, full-path normalization, and reader diagnostic head on samples from $\mathcal{G}_{\text{diag}}$.
8. Record Stage 2 normal reader accuracy as a training-progress diagnostic.
9. Initialize the main classifier from the local auxiliary classifier.
10. Freeze all modules except full-path normalization and the main classifier.
11. Fit full-path normalization and the main classifier on document labels, and select the final checkpoint by validation macro-F1.
12. Reload the selected final checkpoint; recompute writer and reader diagnostics and the route-performance contrast, recording $A_{\text{post}}$, $\Delta A_{\text{cap}}$, $R_{\text{cap}}$, $A_{\text{reader}}^{\text{normal}}$, $A_{\text{reader}}^{\text{shift}}$, $U_{\text{reader}}$, $F_{\text{local}}$, $F_{\text{normal}}$, and $\Delta_{\text{route}}$.
13. Apply the predefined protected-run acceptance criteria.
14. Return the selected model and protocol diagnostics.

Validation acceptance criteria, checkpoint selection, and held-out test locking are specified in Section 5.

### 4.8 Parameter and Computational Analysis

The factorized embedding contains $P_{\text{embed}}$ parameters rather than $Vd$:

$$P_{\text{embed}} = Vr_e + r_e d. \qquad (33)$$

For the configuration evaluated in this study, $V = 50{,}257$, $r_e = 32$, and $d = 64$, giving $P_{\text{embed}} = 50{,}257(32) + 32(64) = 1{,}610{,}272$ embedding parameters. The non-embedding encoder contains 85,152 parameters. The task-model accounting includes two classification heads: the main classifier and the Stage 1 local auxiliary classifier. Each contains $C(d+1) = 65C$ parameters for $d = 64$. The local auxiliary classifier is included in task-model parameter accounting because it is a trainable component of the downstream training architecture, although it is not executed in the final task-inference path. The writer and reader diagnostic heads contain 520 parameters each and are used only by the controlled capability protocol; they are excluded from the task-model count. The resulting task-model parameter count is

$$\begin{aligned} P_{\text{model}} &= P_{\text{embed}} + P_{\text{encoder}} + 2C(d+1) \\ &= 1{,}610{,}272 + 85{,}152 + 2C(65) \\ &= 1{,}695{,}424 + 130C. \end{aligned} \qquad (34)$$

This gives 1,695,684 task-model parameters for IMDB ($C = 2$), 1,695,944 for AG News ($C = 4$), and 1,696,074 for Yelp Review Full ($C = 5$). The instantiated task-inference path excludes the local auxiliary classifier but retains the complete trained encoder. Using $T$, $d$, $r$, and $k$ as asymptotic variables and treating the architecture's fixed channel-expansion ratios as constants, forward inference requires $O(T(kd + d^2))$ operations in the local pathway, $O(T(d^2 + dr + r^2))$ in the writer, and $O(T(dr + r^2))$ in finalized-memory reading. The persistent within-document matrix state requires $O(r^2)$ storage and does not grow with $T$. Training additionally retains token-level activations, including $O(Td)$-scale local representations, together with implementation-dependent autograd state. The writer scan remains sequential across token positions. In contrast, the depthwise local convolution and token-wise finalized-memory reads can be parallelized across positions once $M_T$ is available. Parameter count and persistent-state size, hence, do not, by themselves, determine wall-clock efficiency or total accelerator-memory consumption. Runtime, throughput, and accelerator-memory use are measured empirically under the protocol described in Section 5.

## 5 EXPERIMENTAL PROTOCOLS

The experimental protocol evaluates Protected QL-Memory along three separate dimensions: downstream document-classification performance, preservation of the controlled binding capability during task adaptation, and sensitivity of diagnostic and natural-text evaluation metrics to the selected finalized memory. To prevent evidence from one dimension from substituting for another, capability diagnostics and predefined validation criteria are applied before held-out test evaluation, while finalized-memory interventions are evaluated with all post-intervention parameters and operations fixed. All model-selection, protocol-acceptance, and intervention-based decisions use training or validation data only. The held-out test partitions are reserved for final predictive and efficiency evaluation of the preselected checkpoints.

### 5.1 Datasets and Preprocessing

We evaluate the models on three document-classification datasets accessed through the Hugging Face Datasets ecosystem [22]: stanfordnlp/imdb [19], fancyzhx/ag_news [52], and Yelp/yelp_review_full [52]. Pinned dataset revisions fix the source data used throughout the benchmark. For each dataset, we first index the original training partition and then sample a label-stratified validation subset using seed 42. The validation subsets contain 5,000 IMDB reviews, 7,600 AG News articles, and 10,000 Yelp Review Full reviews. The original test partitions are retained unchanged and excluded from model selection and protocol acceptance. Table 5 summarizes the resulting prepared splits.

Table 5. Source partition sizes and prepared benchmark splits. Percentages in the prepared train, validation, and test columns are relative to the complete prepared corpus for each dataset.

| Dataset | Classes | Source train | Source test | Prepared train | Validation | Prepared test | Prepared corpus total |
|---|---|---|---|---|---|---|---|
| IMDB | 2 | 25,000 | 25,000 | 20,000 (40.00%) | 5,000 (10.00%) | 25,000 (50.00%) | 50,000 |
| AG News | 4 | 120,000 | 7,600 | 112,400 (88.09%) | 7,600 (5.96%) | 7,600 (5.96%) | 127,600 |
| Yelp Review Full | 5 | 650,000 | 50,000 | 640,000 (91.43%) | 10,000 (1.43%) | 50,000 (7.14%) | 700,000 |

No supervised examples from the source training or test partitions are discarded from the prepared corpus. Examples selected for validation are reassigned from the source training partition to the prepared validation split, while the source test partition is retained unchanged as the prepared test split. Accordingly, the prepared corpus total for each dataset equals the sum of its supervised source training and test partitions. Unlabeled partitions and external corpora are not used. Label-stratified sampling preserves the class counts required for exact class balance within each prepared split. Dataset-provided text is passed directly to the tokenizer. The preprocessing pipeline does not apply lowercasing, HTML stripping, stop-word removal, stemming, deduplication, data augmentation, label remapping, or text-based example filtering. After tokenization, the original source columns are removed from the cached tensor representation. All datasets use GPT2TokenizerFast from the Transformers library [48] with the openai-community/gpt2 tokenizer; GPT-2 provides the underlying tokenizer and vocabulary provenance [34]. In the pinned tokenizer artifact, the vocabulary contains 50,257 tokens, and no padding token is assigned. We therefore reuse the end-of-text token, identifier 50,256, for right padding. Token validity is determined exclusively by the explicit attention-derived prefix mask $m$, not by token identity. Reuse of identifier 50,256 so it does not define padding semantically within the model.

Tokenization uses right truncation and maximum-length padding to 512 positions and returns an explicit attention mask. Preprocessing is performed in batches of 512 with one worker. For each cached example, the non-padding length is computed as the sum of the attention mask. The preprocessing cache is materialized at 512 positions to support deterministic artifact reuse. All benchmark models evaluated in this study, however, consume only the first $T = 256$ positions. At benchmark loading, the input identifiers and attention masks are restricted to this prefix, after which effective sequence and non-padding lengths are recomputed. Hence, the empirical claims in this study concern 256-token document prefixes rather than 512-token or unrestricted long-context inputs. Each prepared example is assigned a deterministic identifier of the form dataset-source_split-source_index, linking it to the source dataset, partition, and source index. The preparation artifacts retain the source and prepared split indices, split statistics, dataset and tokenizer provenance, and configuration records required to reconstruct the prepared benchmark. Additional artifact-retention details are provided in Section 5.7.

### 5.2 Baselines and Model Matching

We compare Protected QL-Memory with two compact sequence encoders: a GRU and a Transformer. The GRU [4] contains one recurrent layer with hidden size 112. The Transformer [41] contains two encoder layers with model width 64, four attention heads, and feed-forward width 128. Both controls use the same factorized embedding parameterization [21], configured at rank 32 and width 64, as Protected QL-Memory. The three model families share the tokenizer, prepared examples, batch size, evaluated input length, and paired downstream random seeds. Matching the factorized vocabulary representation reduces the extent to which the common embedding table dominates parameter-count differences and places the sequence encoders at a comparable compact parameter scale. The models are not constrained to exact parameter equality. Their training procedures are also not identical because the capability-protected schedule is part of the evaluated Protected QL-Memory method. RQ2, consequently, compares the complete trained systems under a common downstream benchmark rather than isolating the matrix architecture through a single-component ablation. The GRU provides a gated recurrent reference, whereas the Transformer provides an attention-based reference. The comparison is intended to supply compact recurrent and attention-based control points rather than a state-of-the-art survey of document-classification architectures. Fast-weight, external-memory, and selective state-space systems are discussed in Section 2 as related architectural families but are not included in the locked empirical benchmark. Downstream comparisons are therefore interpreted only with respect to the GRU and Transformer controls evaluated here.

### 5.3 Training Configuration

The downstream experiments use paired random seeds 42, 43, and 44 with deterministic execution enabled. Within each dataset, seed indices are paired across the GRU, Transformer, and Protected QL-Memory runs for downstream comparison. Baseline GRU and Transformer training and Protected QL-Memory Stages 1 through 3 use AdamW [26] with a learning rate of $10^{-3}$, weight decay of $10^{-2}$, batch size 64, dropout 0.1, and gradient clipping at 1.0. No architecture-specific learning-rate schedule is used. The GRU and Transformer controls are each trained for a maximum of seven epochs. Early stopping monitors validation macro-F1 with patience 3 and min_delta=0. All nine GRU runs and all nine Transformer runs complete the full seven-epoch maximum, so early stopping does not terminate any baseline run before the epoch limit. For each run, the checkpoint with the highest validation macro-F1 is stored as checkpoints/best.pt and is used for locked-test evaluation. Stage 0 canonical writer acquisition was configured separately from downstream training. It uses AdamW with a learning rate of $10^{-3}$, diagnostic weight decay of $10^{-4}$, batch size 64, dropout 0.1, and gradient clipping at 1.0. Stage 0 checkpoint progression is governed by controlled-diagnostic accuracy rather than document-classification macro-F1. The controlled binding diagnostic uses key-value pairs and eight value classes. Key identifiers begin at 48,000, value identifiers begin at 48,100, and the terminal-query marker is assigned identifier 48,200. These identifiers are entries in the unchanged GPT-2 vocabulary and are not introduced as additional special tokens.

Canonical writer acquisition follows a fixed sequence-length curriculum. The first 1,000 optimization steps use length 128, the next 1,000 use length 192, and training then continues for at most 6,000 steps at length 256. Diagnostic evaluation is performed every 500 steps, with 0.95 accuracy used as the early acceptance target. Candidate writer checkpoints are subsequently evaluated across the nine dataset-seed transfer conditions defined in Section 5.4. A candidate is eligible for canonical selection only if it attains at least 0.95 controlled binding accuracy in every transfer check. The selected canonical checkpoint is identified in Section 6. Here, the canonical writer denotes the selected Stage 0 writer-interface parameter checkpoint, comprising the shared factorized embedding, pair-context projection, matrix-writer parameters, and writer diagnostic head. The same writer-interface checkpoint is loaded into every protected downstream run. Each input document nevertheless generates its own within-document matrix trajectory and finalized state $M_T$.

For Protected QL-Memory, the downstream seeds vary task-specific initialization and optimization after canonical loading; they do not repeat or resample Stage 0 writer acquisition. The three-seed downstream analysis therefore characterizes variability in downstream adaptation from the fixed selected writer interface rather than variability in writer-capability acquisition. After canonical writer selection, downstream adaptation follows the capability-protected schedule defined in Section 4. Stage 1 runs for four epochs with diagnostic-replay loss weight $\alpha = 0.25$. Stage 2 trains the finalized-memory query, memory-output projection, fusion gate, full-path normalization, and reader diagnostic head for 800 diagnostic steps. After the main classifier is initialized from the selected local auxiliary classifier, all modules except full-path normalization and the main classifier are frozen. These two components are trained in Stage 3 for at most three epochs. The local-route and final full-path checkpoints are selected independently using their respective validation macro-F1 scores. Thus, downstream classification checkpoints are selected by validation macro-F1, whereas Stage 0 writer selection is based on controlled diagnostic accuracy and the transfer criteria defined below.

### 5.4 Capability Diagnostics and Validation Acceptance Criteria

Before downstream adaptation, the canonical writer is evaluated under each combination of the three datasets and three downstream seeds. During optimization, diagnostic distractor tokens are copied from training batches. During diagnostic evaluation, they are copied from validation batches. The controlled key, value, and terminal-query positions then replace the token identifiers at their designated sequence positions. This construction preserves the training-validation split boundary for diagnostic source content while holding the controlled binding structure fixed. Let $A_{\text{canon}}$ denote controlled binding accuracy for a candidate canonical writer-interface checkpoint in a specified dataset-seed validation transfer condition. Canonical selection requires $A_{\text{canon}} \geq 0.95$ in each of the nine transfer checks. Table 6 lists the predefined validation criteria used for canonical selection, downstream entry, and final protected-run acceptance. All thresholds were fixed before the held-out test evaluation.

Table 6. Predefined validation acceptance criteria

| Criterion | Threshold | Interpretation | Protocol rationale | Fixed before test |
|---|---|---|---|---|
| Canonical transfer | $A_{\text{canon}} \geq 0.95$in every transfer check | Controlled binding accuracy of at least 0.95 in each validation transfer condition | Requires high controlled-binding accuracy across all nine dataset-seed validation conditions before canonical selection | Yes |
| Downstream entry | $A_{\text{pre}} \geq 0.90$ | Initial controlled binding accuracy of at least 0.90 | Prevents downstream adaptation from starting from an already degraded writer interface | Yes |

| Criterion | Threshold | Interpretation | Protocol rationale | Fixed before test |
|---|---|---|---|---|
| Capability retention | $R_{\mathrm{cap}} \geq 0.95$ | Post-adaptation diagnostic accuracy is at least 95% of the pre-adaptation score | Limits relative degradation of the measured writer capability to 5% | Yes |
| Reader utility | $U_{\mathrm{reader}} \geq 0.50$ | Normal-memory diagnostic accuracy exceeds cyclic-shift accuracy by at least 0.50 | Requires a substantial diagnostic sensitivity to example-memory correspondence | Yes |
| Local-route tolerance | $\Delta_{\mathrm{route}} \geq -0.01$ | Selected full-path macro-F1 is no more than 0.01 below selected local-route macro-F1 | Limits route-performance degradation after transfer to the fused classifier | Yes |

The downstream-entry score $A_{\mathrm{pre}}$ is the controlled binding accuracy measured immediately before local-language adaptation. After Stage 1, the same diagnostic yields $A_{\mathrm{post}}$. We retain the absolute capability change $\Delta A_{\mathrm{cap}}$ and the relative capability-retention ratio $R_{\mathrm{cap}}$, as defined in Eq. (5). The canonical-transfer and downstream-entry thresholds are both well above the $1/8 = 0.125$ chance accuracy of the eight-way controlled diagnostic. The capability-retention gate permits at most 5% relative degradation from the pre-adaptation score. Reader utility is the controlled intervention contrast $U_{\mathrm{reader}}$ defined in Eq. (6). The normal condition evaluates each example using its own finalized matrix, whereas the shift condition cyclically reassigns finalized matrices across examples. The threshold $U_{\mathrm{reader}} \geq 0.50$ therefore requires at least a 50-percentage-point normal-minus-shift diagnostic accuracy gap. This is an intentionally stringent operational criterion for a large example-memory correspondence effect; it is not derived from a null-hypothesis significance test. For downstream classification, the local-route tolerance is defined by $\Delta_{\mathrm{route}}$ in Eq. (9). The criterion allows the selected full-path checkpoint to fall by at most 0.01 macro-F1 relative to the independently selected local-route checkpoint. These thresholds are study-specific operational criteria fixed before held-out test evaluation. They are not proposed as universal definitions of memory capability, capability preservation, diagnostic access, or downstream task utilization. The continuous measurements are retained for interpretation irrespective of the corresponding binary gate outcome. These measurements include $A_{\mathrm{pre}}$, $A_{\mathrm{post}}$, $\Delta A_{\mathrm{cap}}$, $R_{\mathrm{cap}}$, $A_{\mathrm{reader}}^{\mathrm{normal}}$, $A_{\mathrm{reader}}^{\mathrm{shift}}$, $U_{\mathrm{reader}}$, $F_{\mathrm{local}}$, $\Delta_{\mathrm{route}}$, and the three natural-text macro-F1 intervention contrasts. No minimum natural-text intervention effect is required for protected-run acceptance. The task-performance contrasts $U_{\mathrm{task}}^{\mathrm{shift}}$, $U_{\mathrm{task}}^{\mathrm{zero}}$, and $U_{\mathrm{task}}^{\mathrm{mean}}$are empirical outcomes used to address RQ3 rather than eligibility conditions for locked-test evaluation. This separation avoids selecting protected runs on the downstream memory-sensitivity result that the intervention analysis is intended to measure.

### 5.5 Finalized-Memory Intervention Evaluation

Finalized-memory interventions are used to evaluate whether controlled diagnostic accuracy and natural-text validation macro-F1 are sensitive to the selected finalized matrix. The intervention is applied at the post-write, pre-read interface defined in Section 4.6, with all post-intervention parameters and operations fixed. We evaluate the normal, shift, zero, and mean conditions defined in Eq. (26). Under the normal condition, each example retains its own finalized matrix. The deterministic cyclic-shift condition is the predefined primary correspondence-breaking intervention. Each example receives the preceding example's finalized matrix under cyclic indexing, thereby breaking the original example-memory pairing while preserving the form of an empirically produced matrix. The zero condition removes finalized-matrix content by replacing the matrix with zeros. The mean condition replaces each finalized matrix with the current batch mean, removing sample-specific variation in the selected matrix while retaining shared batch-level matrix content. The local state and finalized-memory query remain sample specific under all intervention conditions. Natural-text intervention analysis is performed on the complete validation partitions without validation subsampling. The validation loader uses batch size 64 and num_workers=0. Validation examples are not shuffled; only the training split is shuffled. Consequently, intervention evaluation follows the fixed prepared-dataset order.

The normal, shift, zero, and mean conditions are evaluated over the same deterministic sequence of validation batches. Batch membership and example order are held fixed across conditions so that the selected memory condition is the intervention-level change. This detail is particularly important for the shift and mean conditions because both are defined relative to the other finalized matrices in the current batch. Controlled reader diagnostics use the first eight validation batches per intervention condition. For each condition, the controlled diagnostic generator is reset to the same seed before evaluation. This preserves the same diagnostic construction and validation-batch sequence across the normal and cyclic-shift reader evaluations. Diagnostic reader sensitivity is measured using controlled binding accuracy. The normal-minus-shift contrast $U_{\mathrm{reader}}$ quantifies diagnostic accuracy sensitivity to example-memory correspondence. Natural-text task-performance sensitivity is assessed using validation macro-F1 under the four memory conditions, yielding $U_{\mathrm{task}}^{\mathrm{shift}}$, $U_{\mathrm{task}}^{\mathrm{zero}}$, and $U_{\mathrm{task}}^{\mathrm{mean}}$. The natural-text intervention results are validation-set analyses rather than independent locked-test estimates. They form part of the predefined experimental protocol and address RQ3 before held-out test evaluation. As defined in Section 3, these contrasts quantify aggregate macro-F1 sensitivity to replacement of the selected finalized state under fixed post-intervention computation. They do not establish invariance of individual predicted labels, logits, or class-probability vectors when macro-F1 remains unchanged.

### 5.6 Evaluation Metrics and Statistical Reporting

We evaluate downstream predictive performance using accuracy, macro precision, macro recall, macro-F1, ROC-AUC [8], confusion matrices, and expected calibration error (ECE) [12]. Macro-F1 was the downstream checkpoint-selection metric. For binary IMDB classification, ROC-AUC is computed from the positive-class probability. For multiclass AG News and Yelp Review Full, ROC-AUC is computed using macro one-versus-rest aggregation. ECE uses 15 equal-width confidence bins over the maximum predicted class probability. Efficiency measurements include total and recurrent parameter counts, parameter-only storage in MiB, epoch time, training throughput, locked-test evaluation-loop latency and throughput, and peak allocated and reserved accelerator memory. Locked-test latency and throughput refer to the synchronized end-to-end evaluation loop, rather than kernel-only model inference. The timed region includes DataLoader iteration, host-to-device transfer, model forward execution, softmax computation, device-to-host copies, and per-example prediction-row construction. Tokenization is excluded because evaluation reads previously prepared cached tensors. Checkpoint loading, DataLoader construction, post-loop metric computation, and artifact serialization are also outside the timed region. The benchmark contains three paired downstream seeds for each dataset. Given this small number of paired observations, the primary statistical presentation is descriptive. We report all seed-level results, the mean, sample standard deviation, and paired model differences for each observed seed. Descriptive 95% $t$-based intervals are also reported but are interpreted cautiously because each interval is estimated from only three observations. Paired standardized effects are reported as Cohen's $d_z$ [20] where their interpretation is meaningful. The empirical conclusions do not rely on binary null-hypothesis significance claims. With only three paired seeds, the smallest attainable exact two-sided Wilcoxon signed-rank $p$-value is 0.25. Supplementary paired $t$-test and exact Wilcoxon descriptors [47] are reported in supplementary evidence, but neither is used as primary evidence for model superiority. The analysis instead emphasizes the magnitude, direction, and consistency of seed-paired differences across the observed runs. Complete locked-test aggregate metrics, run-level confusion matrices and prediction artifacts and the exact locked-test inference measurement procedure are made available in the supplementary evidence.

### 5.7 Reproducibility and Test Locking

Each experimental run stores the resolved configuration, deterministic-execution record, software and hardware environment description, parameter counts, training history, selected and final checkpoints, raw validation probabilities, and the capability and intervention measurements relevant to that stage. Prepared-data artifacts retain deterministic sample identifiers, source and prepared split indices, per-split label counts, source and prepared dataset fingerprints, resolved preparation configurations and their hashes, tokenizer files, and tokenizer-file hashes. Experiments were executed using Python 3.12.13 and PyTorch 2.11.0 [32] with CUDA 12.8 and cuDNN 9.1.9 on an NVIDIA GeForce RTX 5060 Laptop GPU with 8 GB of memory. Deterministic execution was enabled for the paired downstream runs, and the resolved run configurations record the optimizer, learning rate, weight decay, scheduler setting, stage budgets, and seed used for each experiment. Training and validation data are used for optimization, checkpoint selection, capability diagnostics, protocol acceptance, and finalized-memory intervention analysis. The held-out test partitions are excluded from all of these decisions and are not queried until the planned validation runs are complete and the checkpoint-selection and protected-run acceptance decisions have been fixed. After the locked test plan is fixed, each preselected checkpoint was evaluated once on its corresponding held-out test partition. No hyperparameter adjustment, checkpoint reselection, or model update is performed after test evaluation begins. The held-out test partitions are used only for final predictive and efficiency evaluation. Finalized-memory intervention contrasts used for RQ3 and the protected evaluation protocol are computed on validation data and are not re-estimated on the locked test partitions.

## 6 RESULTS

This section reports the empirical results in the order of the predefined evaluation protocol. We first verify protocol completion and locked-test integrity. We then examine whether the controlled writer capability survived downstream adaptation and whether diagnostic performance remained sensitive to example-memory correspondence (RQ1). Next, we compare locked-test predictive quality with the compact GRU and Transformer controls (RQ2), evaluate natural-text macro-F1 sensitivity to finalized-memory intervention (RQ3), and characterize the parameter, evaluation-loop, and accelerator-memory properties of the protected architecture (RQ4). Finally, we examine the direction and variability of seed-paired locked-test differences. Unless otherwise stated, aggregate values are reported as mean ± sample standard deviation over downstream seeds 42, 43, and 44. RQ1 and RQ3 use the predefined validation-side diagnostics and interventions, whereas RQ2 and RQ4 report locked-test results from the preselected checkpoints.

### 6.1 Protocol Integrity and Locked-Test Completion

All 27 ($3\, model\, families \times 3\, datasets \times 3\, seeds$) planned model-dataset-seed runs completed validation-stage training and checkpoint selection, covering three model families, three datasets, and three paired downstream seeds. Each of the nine Protected QL-Memory downstream runs entered adaptation above the predefined $A_{\text{pre}}$ threshold and satisfied the capability-retention, reader-utility, and local-route acceptance criteria before held-out test evaluation. Validation-based checkpoint selection and protocol decisions were fixed before the test partitions were evaluated. A separate artifact-verification pass confirmed completion of all 27 locked-test evaluations and accounted for 743,400 prediction records. Table 7 summarizes protocol completion. Artifact verification reconciled the selected checkpoints, prediction outputs, and suite manifests for all planned evaluations. The complete digest records are retained in the reproducibility manifests.

Table 7. Protocol integrity and locked-test completion

| Protocol item | Verified outcome |
|---|---|
| Planned model-dataset-seed runs | 27/27 |
| Protected downstream runs satisfying acceptance criteria | 9/9 |
| Locked-test evaluations | 27/27 |
| Verified prediction records | 743,400 |

### 6.2 RQ1: Was the Acquired Capability Preserved and Diagnostically Accessible?

The canonical-selection procedure identified the IMDB seed-42 Stage 0 writer-interface checkpoint. Across the nine dataset-seed validation transfer conditions, controlled binding accuracy ranged from 0.9668 to 0.9863 and exceeded the predefined $A_{\text{canon}} \geq 0.95$ criterion in every condition. Across the nine Protected QL-Memory downstream runs, the controlled writer diagnostic showed no post-adaptation degradation. Mean $A_{\text{pre}}$ was 0.9798, whereas $A_{\text{post}}$ reached 1.00 in every run. Consequently, $\Delta A_{\text{cap}}$ was positive for all runs and every $R_{\text{cap}}$ exceeded the predefined 0.95 retention criterion. Table 8 reports the absolute capability change together with the post/pre ratio because $R_{\text{cap}}$ alone does not show whether the underlying diagnostic score increased or decreased.

Table 8. Controlled writer capability and diagnostic reader-access measurements, reported as mean $\pm$ sample SD over seeds 42 to 44.

| Dataset | $\boldsymbol{A_{\text{pre}}}$ | $\boldsymbol{A_{\text{post}}}$ | $\boldsymbol{\Delta A_{\text{cap}}}$ | $\boldsymbol{R_{\text{cap}}}$ | $\boldsymbol{A_{\text{reader}}^{\text{normal}}}$ | $\boldsymbol{A_{\text{reader}}^{\text{shift}}}$ | $\boldsymbol{U_{\text{reader}}}$ |
|---|---|---|---|---|---|---|---|
| AG News | 0.980 ± 0.010 | 1.000 ± 0.000 | 0.020 ± 0.010 | 1.021 ± 0.010 | 0.999 ± 0.001 | 0.131 ± 0.007 | 0.869 ± 0.008 |
| IMDB | 0.980 ± 0.011 | 1.000 ± 0.000 | 0.020 ± 0.011 | 1.021 ± 0.012 | 1.000 ± 0.000 | 0.131 ± 0.007 | 0.869 ± 0.007 |
| Yelp Review Full | 0.980 ± 0.010 | 1.000 ± 0.000 | 0.020 ± 0.010 | 1.021 ± 0.010 | 0.993 ± 0.009 | 0.130 ± 0.007 | 0.863 ± 0.013 |

The reader diagnostic produced a distinct access result. With each example's own finalized matrix, mean reader diagnostic accuracy was 0.9993 on AG News, 1.00 on IMDB, and 0.9928 on Yelp Review Full. Cyclic reassignment reduced mean accuracy to approximately 0.13 on all three datasets, close to the eight-way chance level of 0.125. The resulting $U_{\text{reader}}$ ranged from 0.8626 to 0.8691. Thus, high diagnostic accuracy through the terminal fused representation was strongly sensitive to preservation of the example-memory correspondence. Thus, RQ1 yielded two distinct findings. The measured writer capability was preserved under downstream adaptation, and diagnostic performance through the finalized-memory fused pathway remained strongly sensitive to sample-specific example-memory correspondence. However, neither result establishes natural-text task-performance sensitivity to the selected finalized state; that question is evaluated separately in RQ3.

### 6.3 RQ2: How Did Predictive Quality Compare with the Compact Controls?

Protected QL-Memory was not uniformly the highest-performing model across the three document-classification datasets. Table 9 reports locked-test macro-F1 from the checkpoints selected using validation data. Complete locked-test accuracy, macro precision, macro recall, ROC-AUC, and ECE results are reported in the supplementary evidence.

Table 9. Locked-test macro-F1, reported as mean $\pm$sample SD over the three paired downstream seeds

| Dataset | Protected QL-Memory | GRU | Transformer |
|---|---|---|---|
| AG News | 0.909 ± 0.002 | 0.906 ± 0.001 | 0.910 ± 0.001 |
| IMDB | 0.818 ± 0.003 | 0.828 ± 0.001 | 0.825 ± 0.001 |
| Yelp Review Full | 0.626 ± 0.002 | 0.627 ± 0.003 | 0.602 ± 0.004 |

On AG News, Protected QL-Memory exceeded the GRU by 0.0032 macro-F1 and was 0.0001 below the Transformer. On IMDB, it was 0.0099 below the GRU and 0.0073 below the Transformer. On Yelp Review Full, it was 0.0006 below the GRU and 0.0247 above the Transformer. Accuracy broadly followed the same dataset-dependent ordering. The locked-test results do not support a general predictive advantage for Protected QL-Memory. Its relative position was dataset dependent: the mean AG News macro-F1 differed from the Transformer by 0.0001, the protected model was below both controls on IMDB, and its Yelp Review Full mean was close to the GRU but 0.0247 above the compact Transformer. Validation-leading mean performance did not consistently translate into the highest locked-test mean. Protected QL-Memory had the highest mean validation macro-F1 on AG News and Yelp Review Full, whereas the corresponding locked-test means were narrowly led by the Transformer and GRU, respectively. The changes were small, but they reinforce the distinction between validation-based checkpoint selection and evidence of general predictive superiority.

### 6.4 RQ3: How Sensitive Was Natural-Text Classification Performance to Finalized-Memory Intervention?

#### *6.4.1 Controlled Reader Intervention*

The cyclic-reassignment intervention produced a large reader diagnostic contrast. As reported in Table 8, replacing each example's finalized matrix with the preceding example's matrix under cyclic reassignment reduced diagnostic accuracy by approximately 0.86 to 0.87 while the reader pathway remained fixed. The same pattern occurred in all nine protected runs. Thus, the highly controlled diagnostic accuracy observed under normal memory was strongly sensitive to preservation of the example-memory correspondence.

#### *6.4.2 Natural-Text Memory Interventions*

Natural-text classification exhibited a different intervention response. Table 10 reports validation macro-F1 under the normal, cyclic-shift, zero-memory, and batch-mean conditions. The final column reports the difference between the fused and local classification routes.

Table 10. Validation macro-F1 under normal memory and finalized-memory intervention contrasts. Values are mean ±sample SD over seeds 42 to 44

| Dataset | $F_{\text{normal}}$ | $U_{\text{task}}^{\text{shift}}$ | $U_{\text{task}}^{\text{zero}}$ | $U_{\text{task}}^{\text{mean}}$ | $\Delta_{\text{route}}$ |
|---|---|---|---|---|---|
| AG News | 0.913675 ± 0.002348 | -0.000061 ± 0.000802 | -0.000678 ± 0.000637 | -0.000370 ± 0.000232 | +0.002251 ± 0.000493 |
| IMDB | 0.831251 ± 0.005936 | -0.000197 ± 0.001195 | -0.000006 ± 0.000999 | -0.001992 ± 0.000869 | -0.000933 ± 0.003285 |
| Yelp Review Full | 0.628453 ± 0.004569 | -0.001542 ± 0.000831 | -0.000137 ± 0.000401 | -0.000308 ± 0.000750 | +0.004720 ± 0.007517 |

Note: Positive intervention contrasts indicate lower macro-F1 under intervention than under normal memory; negative values indicate higher intervention macro-F1. $\Delta_{\text{route}}$ is the protocol-level full-path-minus-local contrast and is not an internal memory intervention.

Natural-text classification exhibited a markedly smaller intervention response. Table 10 reports the normal-memory validation macro-F1 together with the normal-minus-intervention contrasts defined in Eqs. (7) and (8). Across the three datasets, the absolute mean $U_{\text{task}}^{\text{shift}}$, $U_{\text{task}}^{\text{zero}}$, and $U_{\text{task}}^{\text{mean}}$values were all below 0.002. The largest absolute mean contrast was 0.001992 on IMDB under batch-mean replacement. The primary cyclic-reassignment intervention, thus, produced a large change in controlled reader diagnostic accuracy but little change in natural-text validation macro-F1. Zero-memory and batch-mean replacement produced the same broad macro-F1 pattern. Under the metric-bounded definition in Section 3, the evaluated natural-text task performance showed low sensitivity to the selected finalized matrix. Fig. 3 shows the finalized-memory intervention results.

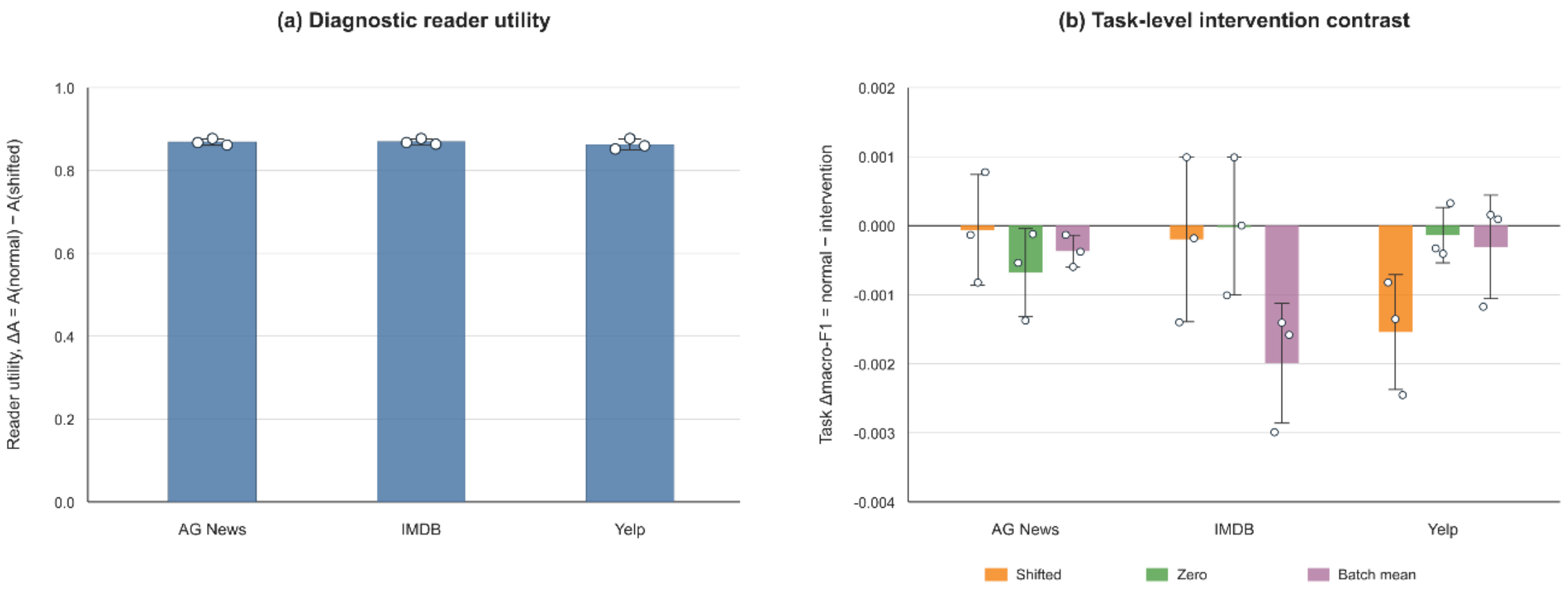

**Fig.** 3 Finalized-memory intervention results. (a) Controlled reader diagnostic accuracy decreases substantially under cyclic reassignment of finalized matrices, showing strong sensitivity to example-memory correspondence. (b) Natural-text validation macro-F1 changes little under cyclic reassignment, zero-memory replacement, or batch-mean replacement. Bars report means, error bars report sample SD, and individual points denote the three downstream seeds. The panels report different outcome measures and are not intended for direct comparison on a common numerical scale

#### *6.4.3 Empirical Separation Between Diagnostic Access and Task-Performance Sensitivity*

The intervention results establish an empirical separation between the controlled memory diagnostics and the natural-text macro-F1 response. The writer retained the measured binding capability, and diagnostic performance through the finalized-memory fused pathway was strongly sensitive to example-memory correspondence. In contrast, replacement of the same finalized-memory input produced absolute mean natural-text macro-F1 contrasts below 0.002 across all evaluated memory conditions and datasets. The protocol-level route contrasts were also small in mean magnitude: $+0.0023$ on AG News, $-0.0009$ on IMDB, and $+0.0047$on Yelp Review Full. Because the local and full-path checkpoints were selected independently, these values are performance contrasts rather than within-model route interventions. The combined results therefore support the narrower conclusion that preserved controlled capability and strong diagnostic sensitivity to sample-specific finalized memory did not imply measurable natural-text macro-F1 sensitivity to that memory under the evaluated interventions. They do not establish the invariance of individual predictions or the relative contribution of the two routes at the prediction level.

### 6.5 RQ4: What Computational Characteristics Accompanied the Protected Architecture?

The three models occupied a narrow total-parameter range of approximately 1.67 to 1.70 million parameters. Protected QL-Memory used approximately 1.1% more parameters than the compact Transformer and 1.5% more than the GRU. The evaluated compact configuration, therefore does not support a parameter-count advantage for the protected architecture. Protected QL-Memory parameter totals follow the active task-model decomposition in Section 4.8. Table 11 reports synchronized locked-test evaluation-loop timing and accelerator-memory measurements at batch size 64 on the same GPU. The timed region includes DataLoader iteration, host-to-device transfer, model forward execution, softmax, device-to-host copies, and per-example prediction-row construction. Tokenization and post-loop metric computation are excluded.

Table 11. Parameter storage and locked-test inference efficiency. Throughput and latency are reported as mean ± sample SD

| Dataset | Model | Parameters | Param. MiB | Test examples/s | Latency ms/example | Peak alloc. MiB | Peak reserved MiB |
|---|---|---|---|---|---|---|---|
| AG News | GRU | 1,670,532 | 6.373 | 612.5 ± 76.2 | 1.650 ± 0.211 | 36.5 | 58.0 |
| | Transformer | 1,677,604 | 6.400 | 759.6 ± 80.3 | 1.327 ± 0.145 | 167.8 | 198.0 |
| | Protected QL-Memory | 1,695,944 | 6.470 | 948.0 ± 59.5 | 1.058 ± 0.069 | 110.9 | 204.0 |
| IMDB | GRU | 1,670,306 | 6.372 | 754.3 ± 171.0 | 1.380 ± 0.359 | 57.7 | 204.0 |
| | Transformer | 1,677,474 | 6.399 | 936.6 ± 71.4 | 1.072 ± 0.085 | 167.8 | 228.0 |
| | Protected QL-Memory | 1,695,684 | 6.469 | 833.6 ± 33.8 | 1.201 ± 0.050 | 110.9 | 232.0 |
| Yelp Review Full | GRU | 1,670,645 | 6.373 | 804.8 ± 22.0 | 1.243 ± 0.034 | 51.8 | 232.0 |
| | Transformer | 1,677,669 | 6.400 | 1,073.3 ± 14.5 | 0.932 ± 0.012 | 167.8 | 232.0 |
| | Protected QL-Memory | 1,696,074 | 6.470 | 818.1 ± 28.3 | 1.223 ± 0.042 | 110.9 | 236.0 |

Evaluation-loop performance was dataset dependent. Protected QL-Memory had the highest measured throughput and lowest per-example latency on AG News, reaching 948.0±59.5 examples/s and 1.058±0.069 ms/example. The Transformer was fastest on IMDB and Yelp Review Full. On Yelp Review Full, Protected QL-Memory, and the GRU had similar measured throughput and latency. These results do not support a universal evaluation-loop speed advantage for any one architecture across the three datasets. Peak allocated accelerator memory for Protected QL-Memory was consistently below the Transformer and above the GRU in the reported measurements. Peak reserved memory was similar to or slightly above the control values. Thus, the evaluated configuration does not establish a universal accelerator-memory advantage for Protected QL-Memory. Stage-level training throughput is retained in the artifacts but is not collapsed into a single cross-model speed ranking. Protected QL-Memory uses heterogeneous optimization stages for local adaptation, controlled reader training, and final classifier fitting, whereas the controls use conventional downstream training loops. Aggregating their stage-level rates would therefore compare different optimization workloads and obscure rather than clarify training cost

### 6.6 Seed-Level Stability and Descriptive Uncertainty

Fig. 4 shows locked-test macro-F1 for each paired downstream seed. On IMDB, Protected QL-Memory was below both controls for all three paired seeds. Its positive difference relative to the Yelp Review Full Transformer was likewise observed for all three seeds. In contrast, the AG News comparisons and the Yelp Review Full comparison with the GRU changed direction across seeds. Table 12 reports the corresponding seed-paired macro-F1 differences, descriptive 95% $t$-based intervals, and Cohen's $d_z$.

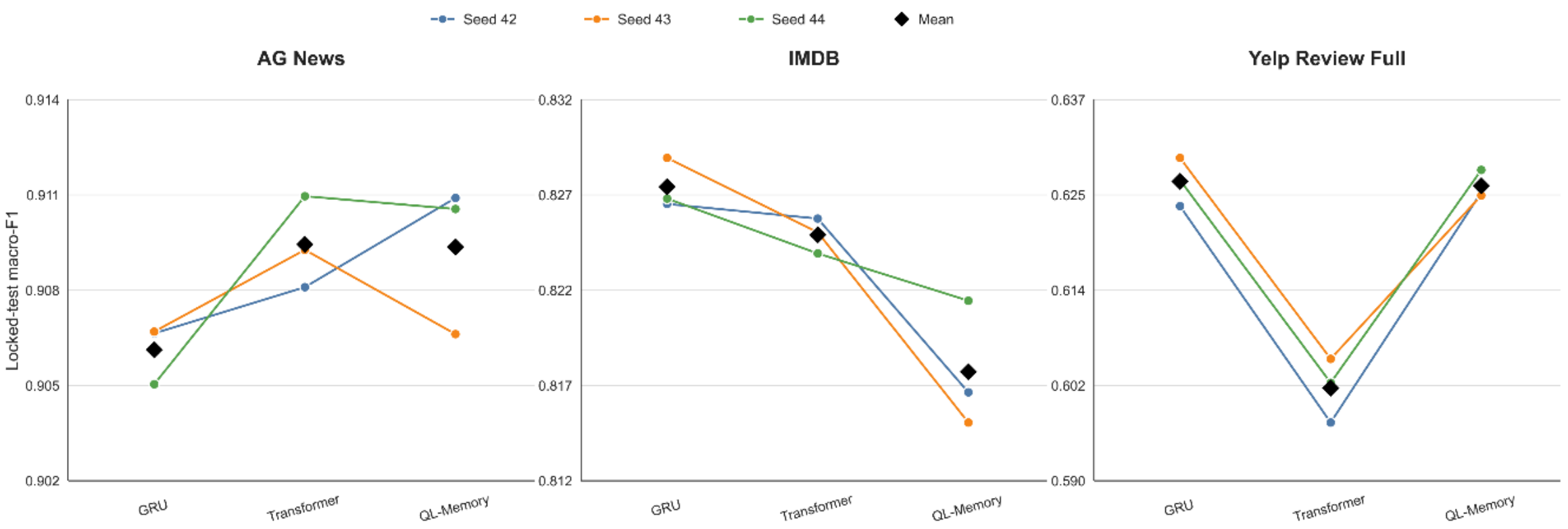

**Fig.** 4 Seed-level locked-test macro-F1. Lines connect results obtained under the same downstream seed across models, and diamonds denote model means. Each panel uses a dataset-specific vertical scale to make the paired differences visible

Table 12. Seed-paired locked-test macro-F1 differences for Protected QL-Memory minus each compact control. Intervals and standardized effects are descriptive because $n = 3$

| Dataset | QL-Memory minus | Seed 42 | Seed 43 | Seed 44 | Mean | Descriptive 95% CI | Cohen's ($d_z$) |
|---|---|---|---|---|---|---|---|
| | GRU | +0.0042 | -0.0001 | +0.0054 | +0.0032 | [-0.0040, +0.0104] | +1.10 |
| AG News | Transformer | +0.0028 | -0.0026 | -0.0004 | -0.0001 | [-0.0068, +0.0067] | -0.03 |
| | GRU | -0.0101 | -0.0141 | -0.0054 | -0.0099 | [-0.0207, +0.0009] | -2.28 |
| IMDB | Transformer | -0.0093 | -0.0102 | -0.0025 | -0.0073 | [-0.0177, +0.0031] | -1.75 |
| | GRU | +0.0017 | -0.0046 | +0.0012 | -0.0006 | [-0.0092, +0.0081] | -0.16 |
| Yelp Review Full | Transformer | +0.0281 | +0.0200 | +0.0260 | +0.0247 | [+0.0142, +0.0352] | +5.84 |

The largest observed mean paired difference was $+0.0247$relative to the Yelp Review Full Transformer, and all three paired differences were positive. On IMDB, the mean differences were $-0.0099$relative to the GRU and $-0.0073$relative to the Transformer, with all three seeds showing the same negative direction in both comparisons. The remaining mean differences were small and either changed direction across seeds or were accompanied by wide descriptive intervals. Given $n = 3$, the intervals and $d_z$values are treated as descriptive summaries rather than population-level evidence.

## 7 DISCUSSION

### 7.1 Preserved Capability Did Not Imply Measurable Task Utilization

The central finding is a separation between preservation of the controlled binding capability, diagnostic sensitivity to sample-specific finalized memory, and measurable downstream task-performance sensitivity to that memory. Protected QL-Memory retained the acquired writer behavior across all nine downstream conditions. Post-adaptation writer diagnostic accuracy reached 1.00 in every run, and diagnostic performance through the terminal fused representation remained strongly sensitive to the pairing between each example and its own finalized matrix. Cyclic reassignment reduced reader diagnostic accuracy to approximately the eight-way chance level, producing mean $U_{\text{reader}}$ contrasts between 0.863 and 0.869. The natural-text interventions produced a different pattern. Cyclic reassignment, zero replacement, and batch-mean replacement changed validation macro-F1 by less than 0.002 in absolute mean magnitude across all three datasets. Thus, the protected schedule retained the measured controlled capability and maintained a fused diagnostic pathway whose performance was strongly sensitive to example-memory correspondence, but these properties did not imply measurable natural-text macro-F1 sensitivity to the selected finalized state. This pattern is not evidence of a capability-preservation failure. The writer diagnostic shows that the measured binding behavior survived downstream adaptation, and the large $U_{\text{reader}}$ contrast shows that controlled diagnostic performance remained sensitive to example-memory correspondence. The separation emerged only when the selected finalized state was intervened on during natural-text classification: aggregate macro-F1 remained nearly unchanged. The unresolved issue is therefore not whether the measured capability survived, but whether and how that preserved capability became aligned with task-relevant evidence strongly enough to alter the evaluated classification metric.

### 7.2 Why Accessible Memory May Not Translate to Task-Performance Sensitivity

The architecture and training objectives provide one plausible explanation for the observed utilization gap. Protected QL-Memory contains a local pathway that contributes directly to document classification, and residual fusion retains the local state as an explicit additive component. The capability-protected schedule constrains degradation of the writer and explicitly trains controlled access through the finalized-memory pathway. The downstream cross-entropy objective, however, contains no term that rewards complementarity between the local and memory routes or requires the task solution to change when the finalized matrix is replaced. Consequently, the optimization objective can, in principle, be minimized without making the memory route essential to the selected task solution. One plausible case is that the local pathway already provides a sufficiently predictive representation, leaving little optimization pressure to exploit memory-derived information. The small $\Delta_{\text{route}}$ values are consistent with this possibility, although they do not identify the internal contribution of either route and do not establish that residual fusion caused the observed low task sensitivity. Because the local and full-path checkpoints were selected independently, $\Delta_{\text{route}}$ is a protocol-level performance contrast rather than a within-model route intervention. The controlled reader objective also differs from the downstream task objective. During Stage 2, the finalized-memory query, memory-output projection, fusion gate, full-path normalization, and reader diagnostic head are trained to support controlled key-value recovery. This stage does not directly train the reader and fusion modules to identify label-relevant evidence in natural text. During Stage 3, all modules except full-path normalization and the main classifier are frozen. Document-label supervision can therefore adjust the final normalized representation and decision boundary, but it cannot reshape the learned finalized-memory query, memory-output projection, or fusion gate toward label-relevant retrieval. Strong controlled binding performance consequently establishes competence on the measured associative operation, not automatic alignment between that operation and the evidence required by the downstream labels.

The dataset-dependent predictive results do not identify which document properties account for the different relative rankings. Protected QL-Memory differed from the Transformer by only 0.0001 mean macro-F1 on AG News, was below both controls on IMDB, and was close to the GRU while above the compact Transformer on Yelp Review Full. The present experiments do not directly measure discourse structure, negation sensitivity, evidence locality, or long-range dependency use and therefore cannot attribute these differences to a specific linguistic or representational factor. The intervention results likewise do not show that the memory route compensated for the protected model's deficits, where it lagged a control. Several alternative explanations remain compatible with the intervention results. Task-relevant information may be represented redundantly in the local and memory routes. The writer may retain the controlled binding operation while encoding natural-text information that is poorly aligned with the classification labels. The reader may achieve high controlled diagnostic accuracy using queries or projections that are not well matched to task-relevant natural-text evidence. In addition, aggregate macro-F1 may conceal intervention-induced changes in individual labels, confidence scores, or class-probability vectors that offset at the dataset level. The present experiments do not distinguish between these possibilities. Direct analysis of what is written to the finalized matrix, what the learned query retrieves, and how the retrieved state changes logits or individual decisions would be required to separate them.

### 7.3 Evaluation Implications for Protected Memory Models

The results motivate a claim-separated view of memory evaluation. For the architecture studied here, the existence of persistent within-document state is a structural property, whereas capability acquisition, capability preservation, diagnostic access, and downstream task-performance sensitivity are distinct learned or behavioral questions. The present experiments associate these questions with architecture and mask-safe state tests, writer diagnostics, pre/post capability measurements, finalized-memory reassignment in the controlled diagnostic, and natural-text macro-F1 intervention contrasts, respectively.

Collapsing these questions into a single performance measure can lead to incomplete conclusions. End-task performance alone would not reveal that the controlled binding capability survived adaptation. Conversely, near-perfect writer diagnostic accuracy and strong reader diagnostic sensitivity to example-memory correspondence would overstate downstream utilization if interpreted as evidence that natural-text classification performance is sensitive to the same finalized state. In the present experiments, either interpretation would miss the observed separation between capability preservation, diagnostic access, and task-performance sensitivity. This separation is the main methodological implication of the study. A memory mechanism should not be characterized as used by a downstream task solely because a controlled diagnostic confirms capability preservation and sample-specific access [7, 36]. Where downstream utilization is part of the architectural claim, the relevant state should also be intervened on after training, while the remaining post-intervention computation is held fixed. Recent activation-patching research similarly emphasizes that intervention construction, effect measurement, and diagnostic design are consequential to mechanistic interpretation [33]. Such a contrast does not estimate a causal effect in the data-generating process. Rather, it provides bounded within-model evidence about the sensitivity of the specified evaluation metric to manipulation of the selected internal state under the tested intervention.

The same caution applies to negative intervention results. The near-zero natural-text contrasts observed here support low validation macro-F1 sensitivity to the selected finalized-memory input under the specified replacements. They do not show that associative memory is generally unnecessary for document classification, that no task-relevant information exists in the matrix, or that individual labels and class-probability vectors are invariant. The conclusion is narrower: under the evaluated architecture, training schedule, 256-token input prefix, and three downstream tasks, replacement of the finalized memory produced little measurable change in validation macro-F1.

### 7.4 Limitations and Threats to Validity

The empirical scope is limited to three document-classification datasets, three downstream seeds, and fixed 256-token document prefixes. The results therefore do not establish behavior at substantially longer sequence lengths, under streaming input, or across broader language-generation and sequence-reasoning tasks. The three-seed design also provides limited precision for uncertainty estimation and little resolution for conventional paired significance tests. The seed-level comparisons should consequently be interpreted through the magnitude, direction, and consistency of the observed differences rather than as population-level significance claims.

The empirical comparison scope is similarly bounded. The locked benchmark includes compact GRU and Transformer controls evaluated with the same prepared data, factorized vocabulary representation, input length, batch size, and paired downstream seeds, but the models are neither exactly parameter matched nor trained with identical procedures. Protected QL-Memory is slightly larger than both controls, and its staged capability-protected training is part of the evaluated method. The benchmark also does not include fast-weight architectures, dedicated external-memory systems, or modern selective state-space models. Consequently, the predictive comparisons are restricted to the two compact controls evaluated here. Moreover, all nine protected downstream runs start from the selected canonical writer-interface checkpoint. Variation across downstream seeds therefore characterizes downstream adaptation and task-training variability but does not quantify variability in writer acquisition or canonical-writer selection. The controlled binding diagnostic is intentionally narrow and may not capture the memory operations required by the three natural-text tasks. The controlled key-value construction tests a specific associative capability, whereas the downstream labels may require different write, query, or aggregation behavior. The diagnostic result should therefore be interpreted as evidence for preservation of the defined controlled capability rather than for general-purpose natural-language memory.

The intervention design also has limitations. Cyclic reassignment and batch-mean replacement are batch dependent. The fixed validation order and identical batch sequence across conditions standardize the within-protocol comparison, but cyclic reassignment may still pair an example with a finalized matrix produced from a semantically similar or same-label document. Batch-mean replacement removes sample-specific variation in the selected matrix while retaining shared batch-level matrix content. Zero replacement removes finalized-memory content entirely but may present the reader with a state unlike those encountered under normal operation. The broadly similar macro-F1 pattern across the three replacement strategies reduces reliance on any one intervention construction but does not eliminate these limitations. Task-performance sensitivity is operationalized using aggregate validation macro-F1. This metric can remain nearly unchanged even when an intervention alters individual predicted labels, confidence values, class probabilities, or decision margins, particularly if positive and negative changes offset across examples. The present study does not report prediction-flip rates or distributional changes in model outputs under finalized-memory intervention. The negative task-level finding should therefore be interpreted as low aggregate macro-F1 sensitivity rather than prediction-level invariance.

Natural-text finalized-memory interventions were evaluated on validation data by design because the intervention hypotheses and contrasts formed part of the predefined validation-side analysis for RQ3 before held-out test evaluation. The $U_{\text{task}}$ contrasts were not protected-run acceptance gates, but they are nevertheless validation-set analyses rather than independent locked-test intervention estimates. The held-out test partitions were reserved for final predictive and efficiency evaluation after model-selection and protocol decisions had been fixed. A confirmatory study focused specifically on task-memory sensitivity should preregister its intervention hypotheses and evaluate them on fresh held-out data.

Finally, the locked-test evaluation-loop timing and accelerator-memory measurements were obtained on one GPU under a fixed synchronized execution procedure. The timed region includes data loading, host-to-device transfer, model forward execution, and prediction-row construction and uses no separate warm-up phase. These values therefore characterize the implemented evaluation pipeline rather than kernel-only inference. The observed results show no universal runtime or accelerator-memory advantage for Protected QL-Memory, but they should not be generalized to other devices, compilers, batch regimes, or optimized kernels without additional benchmarking.

### 7.5 Implications for Future Protected Memory Design

For the evaluated architecture, capability retention alone was insufficient to establish downstream task sensitivity. The results suggest that protected-memory design should treat preservation and task alignment as separate objectives. One direction is to train the finalized-memory query, memory-output projection, or fusion modules with task-conditioned auxiliary objectives after controlled access has been established, while retaining explicit safeguards against writer degradation. Another is to introduce write- or read-side objectives that encourage the finalized memory to encode label-relevant information not already recoverable through the local route. Such designs would need separate intervention tests to determine whether any added task alignment actually increases sensitivity to the selected memory state. The fusion mechanism also warrants closer study. Residual fusion retains the local state as an explicit task route, which can support classification even when the memory-derived state adds little useful signal. This flexibility may also reduce optimization pressure to make the memory route consequential. Future work could examine learned route-selection mechanisms, sparse fusion, or intervention-aware regularization. Such objectives should reward task-relevant sensitivity to memory changes when those changes carry useful evidence rather than indiscriminately forcing the classifier to depend on the matrix.

A second direction is to evaluate protected memory on tasks that more directly require persistent binding, delayed association, or integration of evidence across separated sequence positions. Hybrid evaluations combining natural text with controlled retrieval demands may help determine whether the present utilization gap reflects task sufficiency, objective misalignment, or limitations of the reader interface. Future intervention analyses should also complement aggregate macro-F1 with prediction-flip rates, class-agreement measures, true-class probability changes, and logit-margin or probability-distribution shifts. Repeating canonical writer acquisition across multiple writer-training seeds would further separate variability in capability acquisition from variability introduced during downstream adaptation.

Broader empirical comparisons are also needed. Comparably scaled fast-weight, memory-augmented, and selective state-space controls would help position Protected QL-Memory relative to other mechanisms for persistent sequence state. Longer-context experiments and device-diverse benchmarking would be required before making scaling, systems-efficiency, or deployment claims. Since the test partitions used in the present study have now been evaluated, architectural changes motivated by these findings should not be tuned against the current locked-test outcomes. A task-alignment-oriented redesign requires a new validation protocol and fresh held-out confirmation data.

## 8 CONCLUSIONS

This study introduced Protected QL-Memory to examine a specific question in memory-augmented sequence modeling: whether an acquired associative capability can survive downstream adaptation, remain accessible through a finalized-memory interface, and produce measurable sensitivity in downstream task performance. Across all nine protected downstream runs, the controlled writer capability was preserved after local-language adaptation. Diagnostic performance through the terminal fused representation was strongly sensitive to example-memory correspondence, and cyclic reassignment of finalized matrices reduced reader diagnostic accuracy to approximately the eight-way chance level. These results establish that the measured capability survived adaptation and that sample-specific finalized memory remained diagnostically accessible. The downstream results revealed a different pattern. Protected QL-Memory showed dataset-dependent predictive performance relative to the compact GRU and Transformer controls and did not exhibit a general parameter-count, evaluation-loop runtime, or accelerator-memory advantage. More importantly, natural-text validation macro-F1 changed by less than 0.002 in absolute mean magnitude under cyclic reassignment, zero-memory replacement, and batch-mean replacement across the three datasets. Under the metric-bounded definition adopted in this study, preserved capability and strong diagnostic sensitivity to sample-specific memory did not imply measurable downstream task utilization. Aggregate classification performance remained largely insensitive to replacement of the selected finalized state. Capability preservation is therefore an integrity property of an adapted memory mechanism. It establishes that a previously acquired computation remains available after downstream optimization, but it does not establish that the downstream predictor uses that computation or benefits from it. The principal conclusion is that capability acquisition, capability preservation, diagnostic access, and downstream task-performance sensitivity are distinct empirical properties and should not be inferred from one another. The present findings do not establish that the finalized matrix contained no task-relevant information, that individual predictions were invariant under intervention, or that associative memory is generally unnecessary for document classification. Rather, for the evaluated architecture, staged training protocol, 256-token document prefixes, and three classification tasks, preserving and accessing a measured memory capability was insufficient to make validation macro-F1 materially sensitive to that memory. Future protected-memory designs should therefore treat task alignment as a separate optimization and evaluation problem, with task-conditioned memory objectives and prediction-level intervention analyses assessed on fresh held-out data.

## STATEMENTS AND DECLARATIONS

**Funding:** No funding was received for conducting this study.

**Competing Interests:** The author has no relevant financial or non-financial interests to disclose.

**Author Contributions:** Isaac K. Nti conceived and designed the study, developed the methodology and software, conducted the experiments, analyzed and interpreted the results, and prepared and revised the manuscript. The author read and approved the final manuscript.

**Data Availability:** All datasets used in this study are publicly available. Their respective sources and access details are cited in the manuscript.

**Code Availability:** The code used for the experiments reported in this article is currently maintained in a private GitHub repository and will be made publicly available upon publication.

**Ethics Approval:** Not applicable. This study used publicly available benchmark datasets and did not involve direct interaction with human participants or animals.

**Consent to Participate:** Not applicable.

**Consent for Publication:** Not applicable.